\documentclass[runningheads,a4paper,12pt,twoside,oribibl,envcountsame]{article}
  \newdimen\paravsp  \paravsp=1.3ex 
\usepackage[margin=1in]{geometry}
\newenvironment{keywords}{\centerline{\bf\small Keywords}\begin{quote}\small}{\par\end{quote}\vskip 1ex}
\usepackage[mathletters]{ucs} 
\usepackage[utf8x]{inputenc} 
\usepackage{latexsym,amsmath,amssymb,bm}
\usepackage{hyperref} 
\usepackage{wrapfig}  
\usepackage{graphicx} 
\usepackage[dvipsnames]{xcolor} 

\def\paradot#1{\vspace{\paravsp plus 0.5\paravsp minus 0.5\paravsp}\noindent{\bf\boldmath{#1.}}} 
\def\hrefurl#1{\href{#1}{\rule{0ex}{1.7ex}\color{blue}\underline{\smash{#1}}}} 
\def\nq{\hspace{-1em}}          
\def\fr#1#2{{\textstyle\frac{#1}{#2}}} 
\def\P{\mathbb{P}}                 
\def\E{\mathbb{E}}                 
\def\v{\boldsymbol}             
\def\citep{\cite}
\def\citet{\cite}
\def\eqm{\smash{\stackrel{×}=}}
\def\leqm{\smash{\stackrel{×}{≤}}}
\def\geqm{\smash{\stackrel{×}{≥}}}
\def\eqam{\smash{\stackrel{×}{≈}}}
\def\cD{{\cal D}}                
\def\cH{{\cal H}}                
\def\cS{{\cal S}}                
\def\cX{{\cal X}}                
\def\cY{{\cal Y}}                
\def\vth{{\v{θ}}}                
\def\IE{\text{\sf E}}           
\def\EE{\text{\sf E$\!\!\!\!\;$E}}           
\def\AE{\overline{\text{\sf E}}} 
\def\Var{\mathbb{V}}             
\def\Loss{\text{\sf Loss}}       

\begin{document}

\title{\vspace{-4ex}
\normalsize\sc \vskip 2mm\bf\Large\hrule height5pt \vskip 4mm 
Learning Curve Theory
\vskip 4mm \hrule height2pt
}
\author{{\bf Marcus Hutter}\\[3mm]
  \normalsize DeepMind\\[2mm]
  \normalsize \hrefurl{http://www.hutter1.net/}}
\date{5 February 2021}
\maketitle

\begin{abstract}
Recently a number of empirical ``universal'' scaling law papers have been published, most notably by OpenAI.
`Scaling laws' refers to power-law decreases of training or test error w.r.t.\ more data, larger neural networks, and/or more compute.
In this work we focus on scaling w.r.t.\ data size $n$.
Theoretical understanding of this phenomenon is largely lacking,
except in finite-dimensional models for which error typically decreases with $n^{-1/2}$ or $n^{-1}$, where $n$ is the sample size.
We develop and theoretically analyse the simplest possible (toy) model that can exhibit $n^{-β}$ learning curves for arbitrary power $β>0$, 
and determine whether power laws are universal or depend on the data distribution.
%
\vspace{5ex}\def\contentsname{\centering\normalsize Contents}\setcounter{tocdepth}{1}
{\parskip=-2.7ex\tableofcontents}
\end{abstract}


\newpage
\section{Introduction}\label{sec:Intro}

\paradot{Power laws in large-scale machine learning}
The `mantra' of modern machine learning is `bigger is better'.
The larger and deeper Neural Networks (NNs) are, the more data they are fed, the longer they are trained,
the better they perform. 
Apart from the problem of overfitting \citep{Belkin:18} and the associated recent phenomenon of double-descent \citet{Belkin:19},
this in itself is rather unsurprising.  
But recently `bigger is better' has been experimentally quantified,
most notably by Baidu \citep{Hestness:17} and OpenAI \citep{Henighan:20,Kaplan:20,Hernandez:21}.
They observe that the \emph{error} or \emph{test loss} decreases as a power law,
with the \emph{data size}, with the \emph{model size} (number of NN parameters), 
as well as with the \emph{compute budget} used for training, 
assuming one factor is not ``bottlenecked'' by the other two factors.
If all three factors are increased appropriately in tandem,
the loss has power-law scaling over a very wide range of data/model size and compute budget.

If there is intrinsic noise in the data (or a non-vanishing model mis-specification),
the loss can never reach zero, but at best can converge to the intrinsic entropy of the data (or the intrinsic representation=approximation error).
When we talk about \emph{error}, we mean test loss with this potential offset subtracted, similar to regret in online learning.

\paradot{Ubiquity/universality of power laws}
Power laws have been observed for many problem types (supervised, unsupervised, transfer learning) and 
data types (images, video, text, even math) 
and many NN architectures (Transformers, ConvNets, ...) \citep{Hestness:17,Rosenfeld:19,Henighan:20,Kaplan:20}.
This has led some to the belief that power laws might be universal:
Whatever the problem, data, model, or learning algorithm,
learning curves follow power laws.
To which extent this conjecture is true,
we do not know, since theoretical understanding of this phenomenon is largely lacking.
Below we review some (proto)theory we are aware of.

\paradot{Theory: Scaling with model size}
Consider a function $f:[0;1]^d→ℝ$ which we wish to approximate.
A naive approximation is to discretize the hyper-cube to an $ε$-grid.
This constitutes a model with $m=(1/ε)^d$ parameters,
and if $f$ is $L$-Lipschitz, 
can approximate $f$ to accuracy $L⋅ε=L⋅m^{-1/d}$,
i.e.\ the (absolute) error scales with model size $m$ as a power law with exponent $-1/d$.
More generally, there exist (actually linear) models with $m$ parameters  
that can approximate all functions $f$ whose first $k$ derivatives are bounded to accuracy
$O(m^{-k/d})$ \citep{Mhaskar:96}, again a power law, and without further assumptions, 
no reasonable model can do better \citep{DeVore:89};
see \citet{Pinkus:99} for reformulations and discussions of these results in the context of NNs.
Not being aware of this early theoretical work, 
this scaling law has very recently been empirically verified and extended by \citet{Sharma:20}.
Instead of naively using the input dimension $d$,
they determine and use the (fractal) dimension of the 
data distribution in the penultimate layer of the NN.

\paradot{Theory: Scaling with compute}
Most NNs are trained by some form of stochastic gradient descent, efficiently implemented in the form of back-propagation.
Hence compute is proportional to number of iterations $i$ times batch-size times model size.
So studying the scaling of error with the number of iterations tells us how error scales with compute.
The loss landscape of NNs is highly irregular, which makes theoretical analyses cumbersome at best.
At least asymptotically, the loss is locally convex, 
hence the well-understood stochastic (and online) convex optimization
could be a first (but possibly misleading) path to search for theoretical understanding of scaling with compute.
The error of most stochastic/online optimization algorithms scales as a power law $i^{-1/2}$ or $i^{-1}$ 
for convex functions \citep{Bubeck:15,Hazan:16}.

\paradot{Theory: Scaling with data size}
Even less is theoretically known about scaling with data size.
\citet{Cho:20} and \citet{Hestness:17} consider a very simple Bernoulli model:
Essentially they observe that the Bernoulli parameter can be estimated to accuracy $1/\sqrt{n}$ from $n$ i.i.d\ samples, 
i.e.\ the absolute loss (also) scales with $1/\sqrt{n}$ \citep{Hestness:17} 
and the log-loss or KL-divergence scales with $1/n$ \citep{Cho:20}.
Indeed, the latter holds for any loss, locally quadratic at the minimum, 
so is not at all due to special properties of KL as \citet{Cho:20} suggests.
%
These observations trivially follow from the central limit theorem for virtually any finitely-parameterized model
in the under-parameterized regime of more-data-than-parameters.
This is of course always the case for their Bernoulli model, 
which only has one parameter,
but \emph{not} necessarily for the over-parameterized regime some modern NNs work in.
Anyway, the scaling laws identified by OpenAI et al.\ are $n^{-β}$, 
for various $β<1/2$, which neither the Bernoulli nor any finite-dimensional model can explain.

\paradot{Data size vs iterations vs compute}
Above we have used the fact that compute is (usually in deep learning) proportional to number of learning iterations, 
provided batch and model size are kept fixed.
In addition, 
\begin{itemize}\parskip=0ex\parsep=0ex\itemsep=0ex
\item[(i)] in \emph{online learning}, every data item is used only once, 
hence the size of data used up to iteration $n$ is proportional to $n$.
\item[(ii)] This is also true for \emph{stochastic learning algorithms} for some recent networks, such as GPT-3, trained on massive data sets,
where every data item is used at most once (with high probability).
\item[(iii)]
When generating \emph{artificial data}, it is natural to generate a new data item for each iteration.
\end{itemize}
Hence in all of these 3 settings, the \emph{learning curves}, error-with-data-size, error-with-iterations, and error-with-compute, are scaled versions of each other.
For this reason, scaling of error with iterations, also tells us how error scales with data size and even with compute,
but scaling with model size is different.

\paradot{This work}
In this work we focus on scaling with data size $n$.
As explained above, any reasonable finitely-parameterized model
and reasonable loss function leads to a scaling law $n^{-β}$ with $β=\fr12$ or $β=1$,
but not the observed $β<\fr12$. 
We therefore conjecture that any theoretical explanation of power laws for a variety $β$ 
(beyond 0-1 and absolute error implying $β=\fr12$ and locally-quadratic loss implying $β=1$)
requires real-world data of unbounded complexity,
that is, no finite-dimensional model can ``explain'' all information in the data.

Possible modelling choices are (a) scaling up the model with data,
or (b) consider non-parametric models (e.g.\ kNN or Gaussian processes), 
or (c) a model with (countably-)infinitely-many parameters.
We choose (c) for mathematical simplicity compared to (b),
and because (c) clearly separates scaling with data from scaling with model size, unlike (a).
In future, (a) and (b) should definitely also be pursued,
in particular, since we have no indication that our findings transfer.

Within our toy model, we show that for domains of unbounded complexity,
a large variety of learning curves are possible, including \emph{non}-power-laws.
It is plausible that this remains true for most infinite models.
Real data is often Zipf distributed (e.g.\ the frequency of words in text), which is itself a power law. 
We show that this, in our toy model, implies power law learning curves with ``interesting'' $β$,
though most (even non-Zipf) distributions \emph{also} lead to power laws but with ``uninteresting'' $β$.

\paradot{Contents}
In Section~\ref{sec:Setup} we introduce our setup:
classification with countable ``feature'' space and a memorizing algorithm,
the simplest model and algorithm we could come up with that exhibits interesting/relevant scaling behavior. 
In Section~\ref{sec:Rates} we derive and discuss general expressions for expected learning curves and for various specific data distributions:
finite, Zipf, exponential, and beyond, many but not all lead to power laws.
In Section~\ref{sec:Var} we estimate the uncertainty in empirical learning curves.
We show that the signal-to-noise ratio deteriorates with $n$,
which implies that many (costly) runs need to be averaged in practice to get a smooth learning curve.
On the other hand, the signal-to-noise ratio of the time-averaged learning curves tends to infinity,
hence even a single run suffices for large $n$.
In Section~\ref{sec:Exp} we perform some simple control experiments to confirm and illustrate the theory and claims, 
and the accuracy of the theoretical expressions.
In Section~\ref{sec:Extend} we discuss (potential) extensions of our toy model towards a more comprehensive and realistic theory of scaling laws:
noisy labels, other loss functions, continuous features, models that generalize, and deep learning.
Section~\ref{sec:Disc} concludes with limitations and potential applications.
Appendix~\ref{app:Loss} discusses losses beyond 0-1 loss.
Appendix~\ref{app:Var} contains derivations of the expected error, and in particular exact and approximate expressions for the time-averaged variance.
Appendix~\ref{app:Noise} considers noisy labels.
Appendix~\ref{app:SumInt} derives an approximation of sums by integrals, tailored to our purpose.
Appendix~\ref{app:Notation} lists notation.
Appendix~\ref{app:Figures} contains some mores plots.

\section{Setup}\label{sec:Setup}

We formally introduce our setup, model, algorithm, and loss function in this section:
We consider classification problems with 0-1 loss and countable feature space.
A natural practical example application would be classifying words w.r.t.\ some criterion.
Our toy model is a deterministic classifier for features/words sampled i.i.d.\ w.r.t.\ to some distribution.
Our toy algorithm predicts/recalls the \emph{class} for a new \emph{feature} from a previously observed (\emph{feature},\emph{class}) pair, 
or acts randomly on a novel \emph{feature}.
The probability of an erroneous prediction is hence proportional to the probability of observing a new feature,
which formally is equivalent to the model in \citet{Chao:81}.
The usage and analyses of the model and resulting expressions are totally different though.
While \citet{Chao:81}'s aim is to develop estimators for the probability of discovering a new species from data whatever the unknown true underlying probabilities,
we are interested in the relationship between the true probability distribution of the data and the resulting learning curves, 
i.e.\ the scaling of expected (averaged) error with sample size.
In Appendix~\ref{app:Loss} we show that, within a for our purpose irrelevant multiplicative constant, the results also apply to most other loss functions.

\paradot{The toy model}
The goal of this work is to identify and study the simplest model that is able to exhibit power-law learning curves 
as empirically observed by \citet{Hestness:17,Henighan:20,Kaplan:20} and others.
Consider a classification problem $h∈\cH:=\cX→\cY$, e.g.\ $\cY=\{0,1\}$ for binary classification,
where classifier $h$ is to be learnt from data $\cD_n:=\{(x_1,y_1),...,(x_n,y_n)\}∈(\cX×\cY)^n$. 
For finite $\cX$ and $\cY$, this is a finite model class ($|\cH|<∞$), which, 
as discussed above, can only exhibit a restrictive range of learning curves, 
typically $n^{-1}/n^{-1/2}/e^{-O(n)}$ for locally-quadratic/absolute/0-1 error.
In practice, $\cX$ is often a (feature) vector space $ℝ^d$, 
which can support an infinite model class ($|\cH|=∞$) (e.g.\ NNs)
rich enough to exhibit (at least empirically) $n^{-β}$ scaling 
for many different $β\not\in\{\fr12,1\}$, typically $β\ll 1$.
The smallest potentially suitable $\cX$ would be countable, e.g.\ $ℕ$, which we henceforth assume.
The model class $\cH:=ℕ→\cY$ is uncountable and has infinite VC-dimension, 
hence is not uniformly PAC learnable, but can be learnt non-uniformly.%
Furthermore, for simplicity we assume that data $\cD_n:=\{(i_1,y_1),...,(i_n,y_n)\}≡(i_{1:n},y_{1:n})$ 
with ``feature'' $i_t∈ℕ$ ``labelled'' $y_t$ is noise-free = deterministic,
i.e.\ $y_t=y_{t'}$ if $i_t=i_{t'}$. 
Let $h_0∈\cH$ be the unknown true labelling function.
We discuss relaxations of some of these assumptions later in Section~\ref{sec:Extend},
in particular extension to other loss function in Appendix~\ref{app:Loss} and noisy labels in Appendix~\ref{app:Noise}.
Let features $i_t$ be drawn i.i.d.\ with $\P[i_t=i]=:θ_i≥0$ and (obviously) $∑_{i=1}^∞ θ_i=1$.
The infinite vector $\vth≡(θ_1,θ_2,...)$ characterizes the feature distribution.
The labels are then determined by $y_t=h_0(i_t)$.

\paradot{The toy algorithm}
We consider a simple tabulation learning algorithm $A:ℕ×(ℕ×\cY)^*→\cY$ that stores all past labelled features $\cD_n$ and 
on next feature $i_{n+1}=i$ recalls $y_t$ if $i_t=i$, i.e.\ feature $i$ has appeared in the past, 
or outputs, in its simplest instantiation, \emph{undefined} if $i\not\in i_{1:n}$ i.e.\ is new. Formally:
\begin{align}\label{eq:alg}
  A(i,\cD_n) ~:=~ \left\{ { y_t ~~~~~~\text{if ~~~~$i=i_t$ for some $t≤n$} \atop 
                          \bot ~~~~~\text{else~~~i.e.\ if } i\not\in i_{1:n}~~~~~~~~~~} \right.
\end{align}

\paradot{Error}
Algorithm $A$ only makes an error predicting label $y_{n+1}$ if $i\not\in i_{1:n}$.
We say $A$ makes 1 unit of error in this case. 
Formally, the \emph{(instantaneous) error} $\IE_n$ of algorithm $A$ when predicting $y_{n+1}$ from $\cD_n$ is defined as 
\begin{align*}
  \IE_n ~:=~ [\![i_{n+1}\not\in i_{1:n}]\!]
\end{align*}
The expectation of this w.r.t.\ to the random choice of $\cD_n$ and $i_{n+1}$ gives the \emph{expected (instantaneous) error}
\begin{align}\label{eq:eerr}
  \EE_n ~:=~ \E[\IE_n] ~=~ \P[i_{n+1}\not\in i_{1:n}] ~=~ ∑_{i=1}^∞ θ_i (1-θ_i)^n 
\end{align}
A formal derivation is given in Appendix~\ref{app:Var}, but the result is also intuitive: 
If feature $i$ has not been observed so far (which happens with probability $(1-θ_i)^n$),
and then feature $i$ is observed (which happens with probability $θ_i$), the algorithm makes an error.
$\EE_n$ as a function of $n$ constitutes an (expected) learning curve, which we will henceforth study.
In Appendix~\ref{app:Loss} we show that expression \eqref{eq:eerr} 
remains valid within an irrelevant multiplicative constant 
for most other loss functions.

\section{Expected Learning Curves}\label{sec:Rates}

We now derive theoretical expected learning curves for various underlying data distributions.
We derive exact and approximate, general and specific expressions for the scaling of \emph{expected} error with sample size.
Specifically we consider finite models, which lead to exponential error decay,
and infinite Zipf distributions, which lead to interesting power laws with power $β<1$.
Interestingly even highly skewed data distributions lead to power laws, albeit with ``uninteresting'' power $β=1$.

\paradot{Exponential decay}
In the simplest case of $m$ of the $θ_i$ being equal and the rest being $0$, 
the error $\EE_n=(1-\fr1m)^n≈e^{-n/m}$ decays exponentially with $n$.
This is not too interesting to us, since 
(a) this case corresponds to a finite model (see above),
(b) exponential decay is an ``artifact'' of the deterministic label and discontinuous 0-1 error, and 
(c) will become a power law $1/n$ after time-averaging (Section~\ref{sec:Var}).

\begin{wrapfigure}{r}{0.4\textwidth}
\vspace{-3ex}\includegraphics[width=0.4\textwidth]{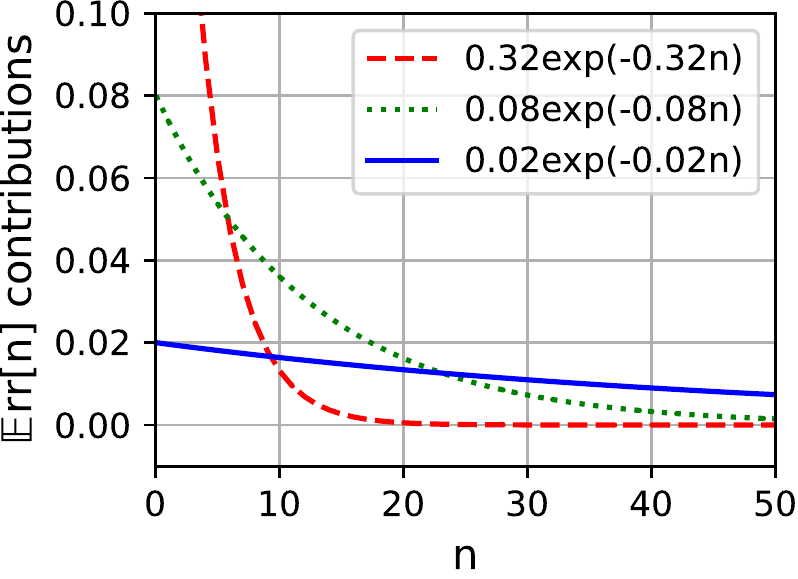}\vspace{-7ex}
\end{wrapfigure}
\paradot{Superposition of exponentials}
Since \eqref{eq:eerr} is invariant under bijective renumbering of features $i∈ℕ$, we can w.l.g.\ assume $θ_1≥θ_2≥θ_3≥...$.
Some $θ$s may be equal. If we group equal $θ$s together into $\bar{\bar{θ}}_j$ with multiplicity $m_j>0$ and define 
$\bar{\bar{ϑ}}_j:=-\ln(1-\bar{\bar{θ}}_j)$, then
\begin{align}\label{eq:seerr}
  \EE_n ~=~ ∑_{j=1}^M m_j \bar{\bar{θ}}_j e^{-n\bar{\bar{ϑ}}_j}
\end{align}
where $M∈ℕ∪\{∞\}$ is the number of \emph{different} $θ_i>0$. 
This is a superposition of exponentials in $n$ (note that $∑_{j=1}^M m_j \bar{\bar{θ}}_j=1$) with different decay rates $\bar{\bar{ϑ}}_j$.
If different $\bar{\bar{θ}}_j$ have widely different magnitudes and/or for suitable multiplicities $m_j$,
the sum will be dominated by different terms at different ``times'' $n$.
So there will be different phases of exponential decay,
starting with fast decay $e^{-n\bar{\bar{ϑ}}_1}$ for small $n$,
taken over by slower decay $e^{-n\bar{\bar{ϑ}}_2}$ for larger $n$, 
and $e^{-n\bar{\bar{ϑ}}_3}$ for even larger $n$, etc.\
though some terms may never (exclusively) dominate,
or phases may be unidentifiably muddled together (see figure above).
In any case, if $M=∞$, the dominant terms shift indefinitely to ever smaller $θ$ for ever larger $n$.
For $M<∞$ eventually $e^{-n\bar{\bar{ϑ}}_M}$ for the smallest $\bar{\bar{ϑ}}$ will dominate $\EE_n$.
The same caveats (a)-(c) apply as for $M=1$ in the previous paragraph.

\paradot{Approximations}
First, in our subsequent analysis we (can) approximate $(1-θ_i)^n=:e^{-nϑ_i}≈e^{-nθ_i}$, justified as follows:
(i) For $nθ_i\ll 1$ this is an excellent approximation.
(ii) For $θ_i\ll 1$, $ϑ_i≈θ_i$, while numerically $e^{-nϑ_i}/e^{-nθ_i}\not≈1$ for $nθ_i\gg 1$,
but the exponential scaling of $e^{-nϑ_i}$ and $e^{-nθ_i}$ we care about is sufficiently similar.
(iii) There can only be a finite number of $θ_i\not\ll 1$,
say, $θ_i$ for $i≤i_0$ are not small, 
then already for moderately large $n\gg 1/θ_{i_0}$, all features $i≤i_0$ are observed with high probability 
and hence do not contribute (much) to the expected error
(formally $e^{-nθ_{_i}}\ll 1$ for $i≤i_0$)
hence can safely be ignored in any asymptotic analysis.

Second, let $f:ℝ→ℝ$ be a smooth and monotone decreasing interpolation of $\vth:ℕ→ℝ$, 
i.e.\ $f(i):=θ_i$ and $f'(x)≤0$. We can then approximate the error as follows:
\begin{align}\label{eq:errapprox}
  \EE_n ~&=~ ∑_{i=1}^∞ f(i) (1-f(i))^n 
  ~≈~ \int_1^∞ f(x)e^{-nf(x)}dx \nonumber\\
  ~&=~ \int_0^{θ_1} {u e^{-nu}du\over|f'(f^{-1}(u))|}
  ~\eqam~ {1\over n^2|f'(f^{-1}(\fr1n))|}
\end{align}
The first $≈$ uses the two approximations introduced above.
The equality follows from a reparametrization $u=f(x)$ and $f(1)=θ_1$ and $f(∞)=0$ and $dx=du/f'(x)$ and $f'<0$.
The numerator $u e^{-nu}$ is maximal and (strongly) concentrated around $u=1/n$, hence $u≈1/n$ gives most of the integral's contribution.
Therefore replacing $u$ by $1/n$ in the denominator can be a reasonable approximation.
The last $≈$ follows from this and $∫_0^{θ_1}u e^{-nu}du≈∫_0^∞ u e^{-nu}du=1/n^2$ for $nθ_1\gg 1$.

Intuitively, the expected error \eqref{eq:eerr} is dominated by samples $i_0$ for which $θ_{i_0}≈\fr1n$.
Estimating the number of such $i_0$ multiplied by $θ_{i_0}(1-θ_{i_0})^n≈θ_{i_0}e^{-1}$ leads to approximation \eqref{eq:errapprox}.
In Appendix~\ref{app:Var} we show that the approximation error of the integral representation is bounded by $1/en+o(1/n)$.

\paradot{Zipf-distributed data}
Empirically many data have been observed to have a power-law distribution, called Zipf distribution in this context, that is,
for a countable domain the frequency of the $i$th most frequent item is approximately proportional to $i^{-(α+1)}$ for some $α>0$.
In our model this will be the case if $θ_i\propto i^{-(α+1)}$, so let $u=f(x)=α⋅x^{-(α+1)}$. 
This implies $x=f^{-1}(u)=(u/α)^{-1/(1+α)}$ and $f'(x)=-α(α+1)x^{-(α+2)}=-α(α+1)(u/α)^{(α+2)/(α+1)}$, hence
\begin{align*}
  &\EE_n ~≈~ {1\over n^2|f'(f^{-1}(\fr1n))|} ~\eqm~ n^{-β},~~~\text{where}~~~β:={α\over 1+α}
\end{align*}
That is, Zipf-distributed data (with power $α+1$) lead to a power-law learning curve (with power $β={α\over 1+α}<1$).
The more accurate integral representation leads to the same power law but with correct coefficient
$\EE_n≈c_α n^{-β}$ with $c_α=α^{1/(α+1)}Γ({α\over 1+α})/(α+1)$.
$c_1=\fr12\sqrt\pi~\dot=~0.886$ and $c_{0.1}=1.177$ in excellent agreement with the fit curves in Figure~\ref{fig:PowZipf}.

\paradot{Exponentially-distributed data}
An exponential data distribution $θ_i\propto e^{-γi}$ is more skewed than any power law.
For $u=f(x)=γ⋅e^{-γx}$ we have $x=f^{-1}(u)=\fr1{γ}\ln\fr{γ}u$ and $f'(x)=-γ^2e^{-γx}=-γu$, 
hence both approximations in \eqref{eq:errapprox} give $1/γn$.
A rigorous upper bound $\EE_n≤(\fr1e+\fr1{γ})\fr1n$ follows from \eqref{eq:eerrapproxA} in Appendix~\ref{app:Var},
and a rigorous lower bound $\EE_n\geqm\fr1n$ from the next paragraph.
So even an exponential data distribution leads to a power law learning curve,
though the exponent $1$ is much larger than observed in (most) experiments,
which hints at that data are not exponentially distributed, 
assuming this toy model has any real-world relevance.

\paradot{Beyond exponentially-distributed data}
For (quite unrealistic) decay faster than exponentially, e.g.\ $θ_i\propto e^{-γi^2}$, the approximations \eqref{eq:errapprox}
are too crude, but somewhat surprisingly we \emph{always} get a (sort of) power law as long as $θ_i>0$ for infinitely many $i$.
First, the previous paragraph implies that $\EE_n\leqm n^{-1}$ for any $θ_i\leqm e^{-γi}$ for any $γ>0$,
i.e.\ the error decreases at least with $n^{-1}$ if the $i$th item has at most exponentially small probability in $i$.
For a (partial) converse, define $n_i:=\lceil 1/ϑ_i\rceil≤1/ϑ_i+1$.
Plugging $ϑ_i:=-\ln(1-θ_i)≥θ_i$ and $2θ_i≥ϑ_i≥1/n_i$ for $θ_i<0.79$ into $\EE_n≥θ_i(1-θ_i)^n=θ_i e^{-nϑ_i}$ we get
\begin{align*}
  \EE_{n_i} ~≥~ θ_i e^{-n_i ϑ_i} ~≥~ \fr1{2n_i}e^{-(1/ϑ_i+1)ϑ_i} ~≥~ \fr1{2n_i e^2} ~\eqm~ n_i^{-1}
\end{align*}
Hence, if $θ_i>0$ for infinitely many $i$, then there are infinitely many $n$ for which 
$\EE_n\geqm n^{-1}$. For $θ_i$ going to zero exponentially or slower,
the spacing between $n_{i+1}$ and $n_i$ has bounded ratio $n_{i+1}/n_i≤e^γ$, 
which implies $\EE_n\geqm n^{-1}$ for all $n$.
For faster decaying $θ_i$, e.g.\ $θ_i=e^{-γi^2}$ this is no longer the case.
So in some weak sense, power law learning curves are universal,
but it's mostly $1/n$, so not useful to explain observed power laws.

\section{Learning Curve Variance}\label{sec:Var}

So far we have considered \emph{expected} learning curves.
This corresponds to averaging infinitely many experimental runs.
In practice, only finitely many runs are possible, sometimes as few as 5 or even 1.
In the following we consider the variance $\Var_n$ of the instantaneous error $\IE_n=[\![i_{n+1}\not\in i_{1:n}]\!]$ as a function of $n$.
The standard error when averaging $k$ runs is then $\sqrt{\Var_n/k}$.
The question we care most about here is whether (for large $n$) this is small or large compared to $\EE_n$,
because this determines whether learning curves (for small $k$) are smooth or look random,
and how large $k$ suffices for a good signal-to-noise ratio.
We also consider time-average expected learning curves and their variance, which are much smoother.
Note that cumulative errors are like attenuated drifting random walks,
a property time-average learning curves qualitatively inherit (red curve in Figure~\ref{fig:zipf} right).

\paradot{Instantaneous Variance}
$\IE_n∈\{0,1\}$, hence $\IE_n^2=\IE_n$, hence
\begin{align*}
  \E[\IE_n^2] ~&=~ \E[\IE_n] ~=~ \sum_{i=1}^∞ θ_i(1-θ_i)^n ~=:~ μ_n, ~~~\text{hence} \\
  \Var[\IE_n] ~&=~ \E[\IE_n^2]-\E[\IE_n]^2 ~=~ μ_n(1-μ_n)
\end{align*}
Since $μ_n→0$ for $n→∞$, the standard deviation 
\begin{align*}
  σ_n ~:=~ \sqrt{\Var[\IE_n]} ~=~ \sqrt{μ_n(1-μ_n)} ~≈~ \sqrt{μ_n} ~~\gg~~ μ_n ~=~ \EE_n
\end{align*}
That is, the standard deviation is much larger than then mean for large $n$.
Indeed, for a single run, there is no proper learning \emph{curve} at all, 
since $\IE_n∈\{0,1\}$ (see Figures~\ref{fig:unifA}\&\ref{fig:zipfA} top left).
In order to get a good signal-to-noise ratio one would need to average a large 
(and indeed increasing with $n$) number $k\gg 1/μ_n$ of runs
(see Figures~\ref{fig:unif}\&\ref{fig:unifA}\&\ref{fig:zipfA}).

\paradot{Time-averaged Mean and Variance}
In practice, beyond averaging over runs, 
other averages are performed to reduce noise.%
One alternative is to report the time-averaged error $\AE$, 
rather than the instantaneous error $\IE$.
We can calculate its mean and variance as follows
\begin{align}
  \AE_N ~:&=~ {1\over N}\sum_{n=0}^{N-1}\IE_n \label{eq:ae}\\
  \E[\AE_N] ~&=~ {1\over N}\sum_{n=0}^{N-1}\E[\IE_n] 
  ~=~ {1\over N}\sum_{i=1}^∞ θ_i\sum_{n=0}^{N-1}(1-θ_i)^n
  ~\stackrel{(a)}=~ {1\over N}\sum_{i=1}^∞ 1-(1-θ_i)^N \label{eq:eae}\\
  \E[\AE_N^2] ~&\stackrel{(b)}=~ {1\over N^2}\sum_{i=1}^∞[1-(1-θ_i)^N] ~+~ {1\over N^2}\sum_{i≠j}[1-(1-θ_i)^N-(1-θ_j)^N+(1-θ_i-θ_j)^N] \nonumber \\
  \Var[\AE_N] ~&\stackrel{(c)}=~ {1\over N^2}\sum_{i=1}^∞(1-θ_i)^N[1-(1-θ_i)^N] ~-~ {1\over N^2}\sum_{i≠j}[(1-θ_i)^N(1-θ_j)^N-(1-θ_i-θ_j)^N] \label{eq:vae}
\end{align}
where $(a)$ is simple algebra, 
$(b)$ follows from inserting the definition of $\AE_N$ and some rather tedious algebraic manipulations (see Appendix~\ref{app:Var}), and 
$(c)$ from inserting $(a)$ and $(b)$ into the definition of variance and simple algebraic manipulation.
We now revisit the exponential and Zipf case studied earlier, 
after a trivial but note-worthy observation.

\paradot{Case \boldmath $θ_i=[\![i=1]\!]$}
In this case $i_n=1~∀n$, hence $\IE_0=1$ and $\IE_n=0~∀n≥1$ and $\Var[\IE_n]=0~∀n$.
This is the fastest any error can decay, $0$ after $1$ observation, 
hence the fastest any time-averaged error can decay is $\AE_N=1/N$.
This means, for any learning problem that has instantaneous error decaying faster than $1/N$,
one should report the instantaneous error rather than the slower decaying and hence much larger time-averaged error.
Most problems of interest in Deep Learning have much slower learning curves though,
and for those, the time-averaged and the instantaneous error have the \emph{same} decay rate,
but the time-averaged has lower variance, so is the preferred one to plot or report.

\paradot{Case \boldmath $θ_i=\fr1m[\![i≤m]\!]$}
In this case, while $\EE_n=(1-\fr1m)^n≈e^{-n/m}$ decays exponentially, 
the average quantities decay with $1/N$ (or $1/N^2$):
\begin{align*}
  \E[\AE_N] ~&=~ \fr{m}{N}[1-(1-\fr1m)^N] ~\longrightarrow~ \fr{m}{N} ~~~\text{for}~~~ N→∞ \\
  \Var[\AE_N] ~&=~ \fr{m^2}{N^2}[\fr1m(1-\fr1m)^N+(1-\fr1m)(1-\fr2m)^N-(1-\fr1m)^{2N}] \\
  ~&→~ \fr{m}{N^2}[e^{-N/m}-e^{-2N/m}] ~→~ \fr{m}{N^2}e^{-N/m}~~~\text{for}~~~N→∞ \\
  σ[\AE_N] ~&≈~ \fr{\sqrt{m}}{N}e^{-N/2m} ~~\ll~~ \fr{m}{N} ~≈~ \E[\AE_N] ~~~\text{for}~~~N\gg m 
\end{align*}
The expressions for the mean and variance follow from the general expression \eqref{eq:eae} and \eqref{eq:vae} above, 
by inserting $θ_i=\fr1m$ for $i≤m$ and noting that $i>m$, for which $θ_i=0$, give no contribution, hence $∑_i 1=m$ and $∑_{i≠j}1=m(m-1)$,
and rearranging terms.
The conclusion most interesting to us is that the standard deviation is (much) smaller than the mean for $N$ (much) larger than $m$,
so the time-averaged learning curves have a much better signal-to-noise ratio; see Figure~\ref{fig:unif}.
Also $σ[\AE_N]≈N^{-1/2}\ll 1≈\E[\AE_N]$ for $m\gg N$. 
The intuition is easy: For $m\gg N$, at ever $n<N$, a new $i_{n+1}$ is observed, i.e.\ $\IE_n=1$ w.h.p.
For $N\gg m$, all $m$ errors have been made, i.e.\ $\AE_N=\fr{m}{N}$ w.h.p, i.e.\ in both cases the variance is small.
Only for $m≈N$ is there sizeable uncertainty in $\IE_N$.
The situation is similar for the most interesting Zipf case:

\paradot{Case \boldmath $θ_i\propto i^{-(α+1)}$}
Recall that for Zipf-distribution $θ_i\propto i^{-(α+1)}$, the expected error followed power law
$\EE_n≈c_α n^{-β}$, where $0<β={α\over 1+α}<1$. The time-averaged error
\begin{align*}
  &\E[\AE_N] ~≈~ {1\over N}\sum_{n=1}^N\E[\IE_n] 
  ~≈~ {c_α\over N}\int_0^N n^{-β}dn 
  ~=~ c'_α N^{-β} ~~~\text{with}~~~c'_α:=\frac{c_α}{1-β}=α^{\frac1{1+α}}Γ(\fr{α}{1+α})
\end{align*}
follows the same power law with the emph{same} exponent $β$, 
which is a generic property as foreshadowed earlier. 
As for the variance, we show in Appendix~\ref{app:Var} that
\begin{align*}
  \E[\AE_N^2] ~&\eqam~ \E[\AE_N]^2,
  ~~~~~ \Var[\AE_N] ~\eqam~~ N^{-{1+2α\over 1+α}}, ~~~\text{hence} \\
  σ[\AE_N] ~&\eqam~ N^{-{1/2+α\over 1+α}} 
  ~~\ll~~ N^{-{α\over 1+α}} ~\eqam~ \E[\AE_N]
\end{align*}
\begin{align*}
  \text{In particular, the signal-to-noise ratio is} 
   ~~ {σ[\AE_N]\over\E[\AE_N]} ~\eqam~ N^{-1/(2+2α)}
\end{align*}
That is, the standard deviation is much smaller than then mean.
A single run suffices to get a good (and excellent for $n\gtrsim 500$) signal-to-noise ratio for the averaged 
and cumulative error 
(see Figures~\ref{fig:PowZipf}\&\ref{fig:TextZipf}\&\ref{fig:TextZipfA} (right) and Figures~\ref{fig:unifA}\&\ref{fig:zipfA} (top left)).
Still, the infinite Zipf model leads to a more noisy learning curve than the finite uniform model.
Intuitively, for every $N$, new $i_{N+1}\not\in i_{1:N}$ have small but sufficient chance, contributing to the error and variance,
decreasing exponentially in the uniform model,
but only as a power law in the Zipf model.

\paradot{\boldmath General $\vth$ Case}
One can show that the signal-to-noise ratio for the time-averaged error improves with $N$ in general for any choice of $\vth$.
First note that the argument of $∑_{i≠j}[...]$ in \eqref{eq:vae} is non-negative,
hence the variance is upper-bounded by the first $∑_i$. Using $1-(1-θ_i)^N≤θ_i N$, we get $\Var[\AE_N]≤\fr1N\EE_N$, so the
\begin{align*}
  \text{signal-to-noise ratio is} ~~~
  {σ[\AE_N]\over\E[\AE_N]} ~≤~ {\sqrt{\smash{\fr1N}\EE_N}\over\E[\AE_N]} ~=~ {\sqrt{N\smash{\EE_N}}\over\sum_{n=0}^{N-1}\EE_n} ~→~ 0 ~~\text{for}~~ N→∞
\end{align*}
To prove the limit $→0$ we have to distinguish two cases: First note that $\EE_n$ is monotone decreasing ($\EE_n\!\!\!\searrow$).
(i) For bounded total error $∑_{n=0}^∞\EE_n≤c$ (e.g.\ exponential error decay in finite models), $\EE_n\!\!\!\searrow$ implies $\EE_N=o(1/N)$,
which implies that the numerator tends to 0; the denominator is lower-bound by $\EE_0=1$.
(ii) For unbounded total error $∑_{n=0}^{N-1}\EE_n→∞$ (most infinite models, e.g.\ Zipf and even exponential $θ_i$), we factor the denominator as
$\sum_{n=0}^{N-1}\EE_n≡\sqrt{\smash{Σ_{n=0}^{N-1}}\EE_n}⋅\sqrt{\smash{Σ_{n=0}^{N-1}}\EE_n}$ and lower-bound one term by $\sum_{n=0}^{N-1}\EE_n≥N\EE_N$,
which is true since $\EE_n\!\!\!\searrow$.

\section{Experiments}\label{sec:Exp}

\begin{figure*}[ptbh!]
\begin{center}
\includegraphics[width=0.49\textwidth]{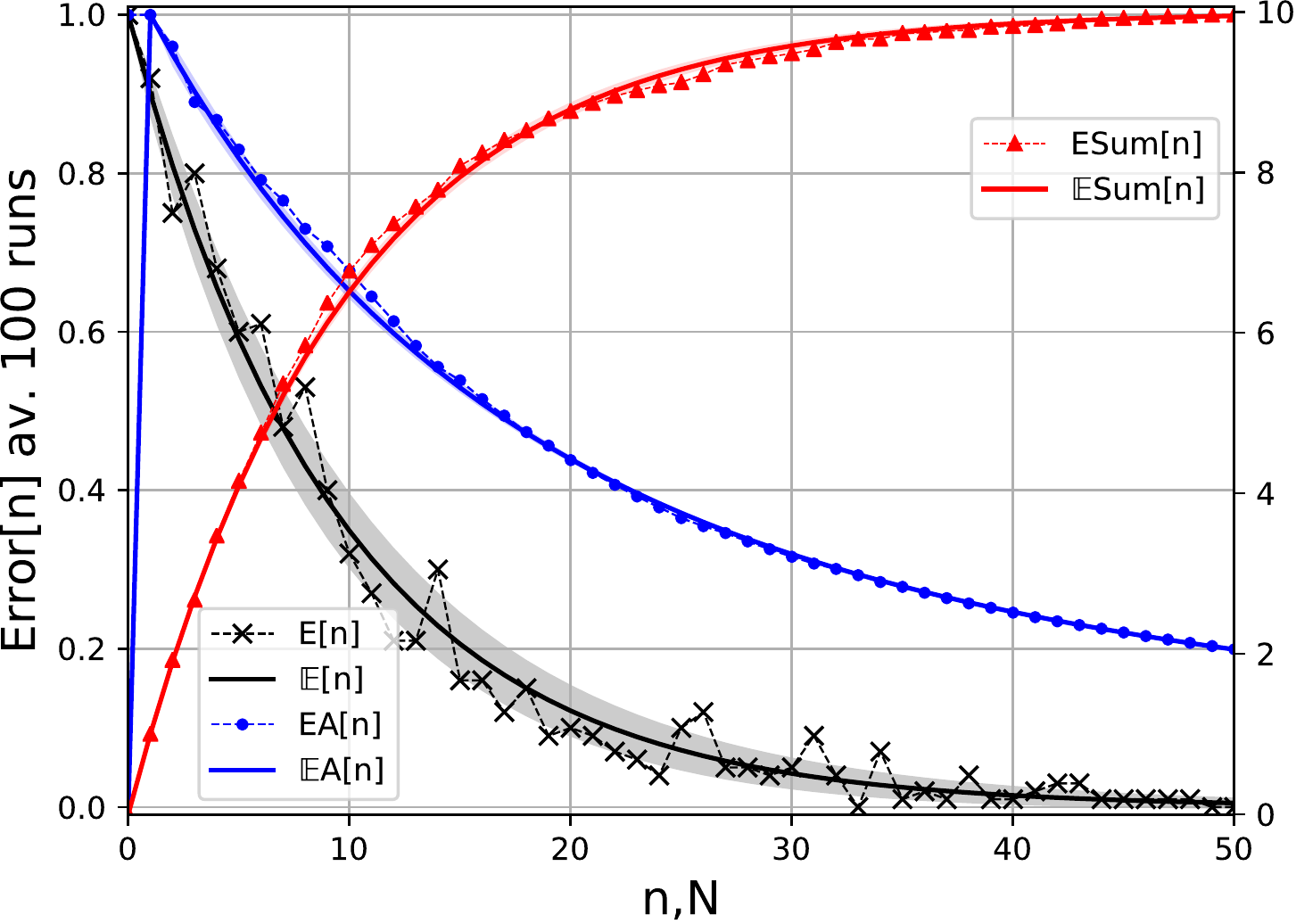}~~%
\includegraphics[width=0.49\textwidth]{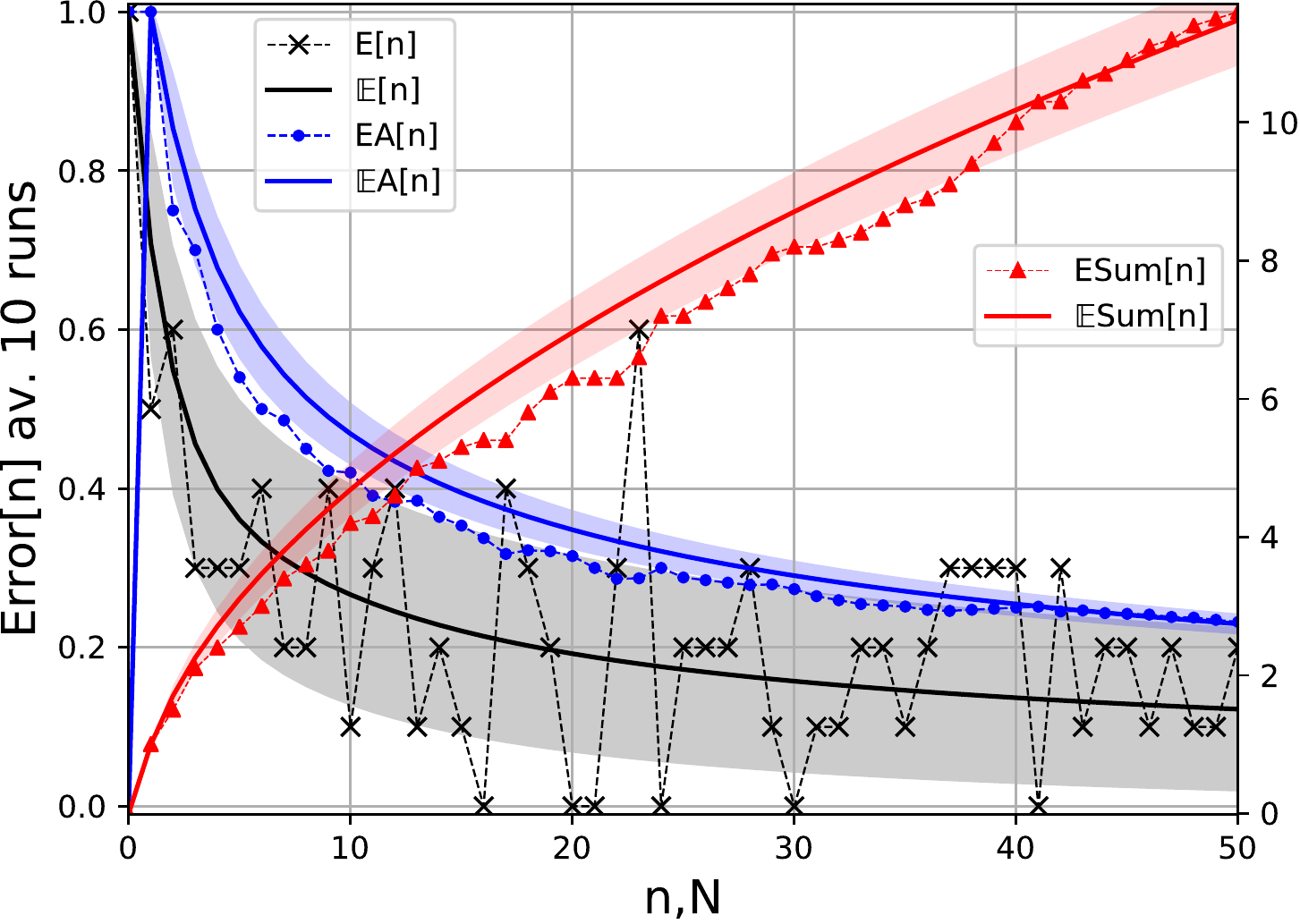}\vspace*{-2ex}
\caption{{(\bf Learning Curves)} \emph{(left)} 
for uniform data distribution $\P[i_n=i]=θ_i=\fr1m$ for $i≤m=10$ averaged over $k=100$ runs.
\emph{(right)} for Zipf-distributed data $\P[i_n=i]=θ_i\propto i^{-(α+1)}$ for $α=1$ averaged over $k=10$ runs. 
See Figures~\ref{fig:unifA} and \ref{fig:zipfA} for more plots for $k=1,10,100,1000$.
}\label{fig:unif}\label{fig:zipf}\vspace*{-2ex}
\end{center}
\end{figure*}

\begin{figure*}[ptbh!]
\begin{center}
\includegraphics[width=0.49\textwidth]{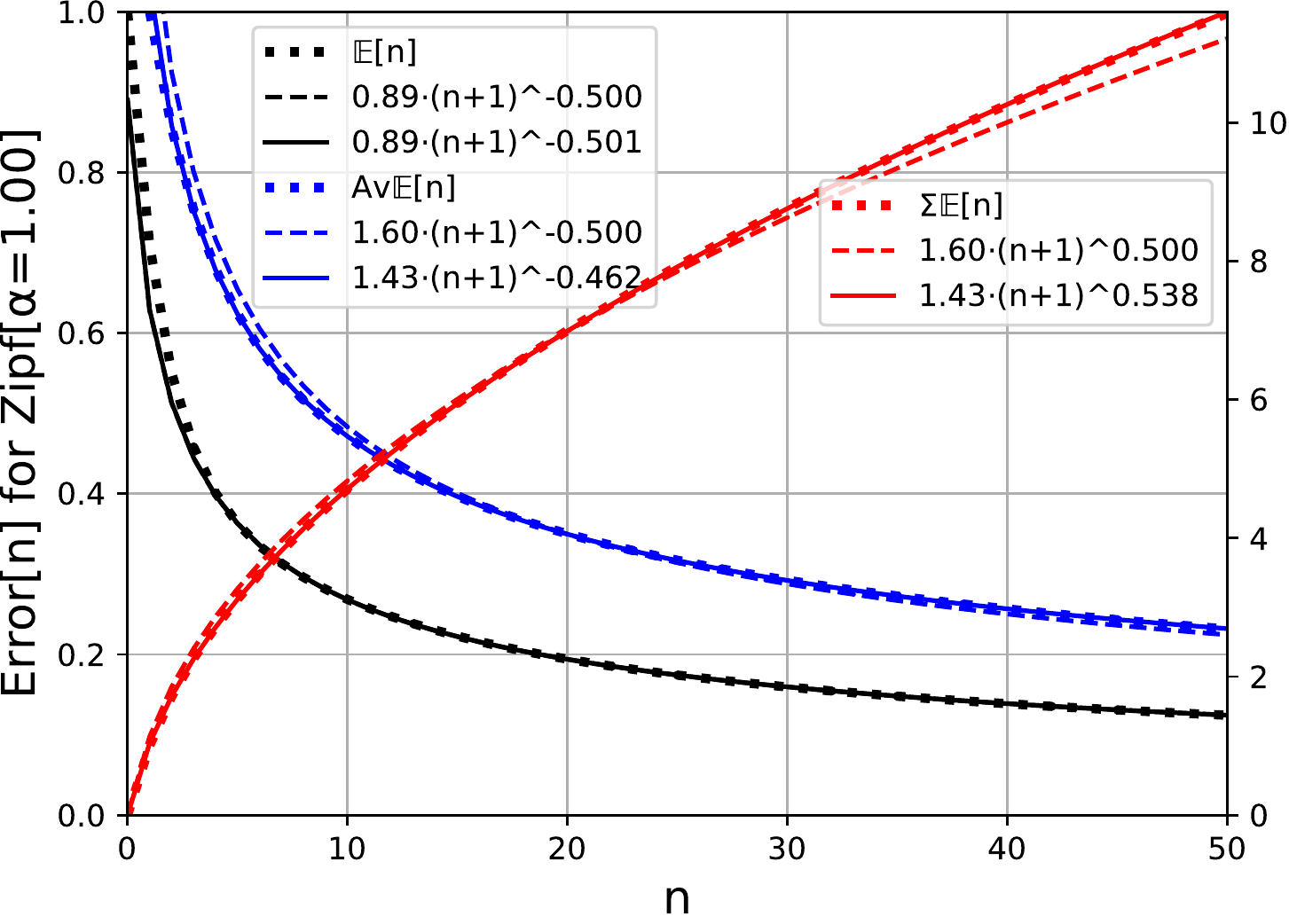}~~%
\includegraphics[width=0.49\textwidth]{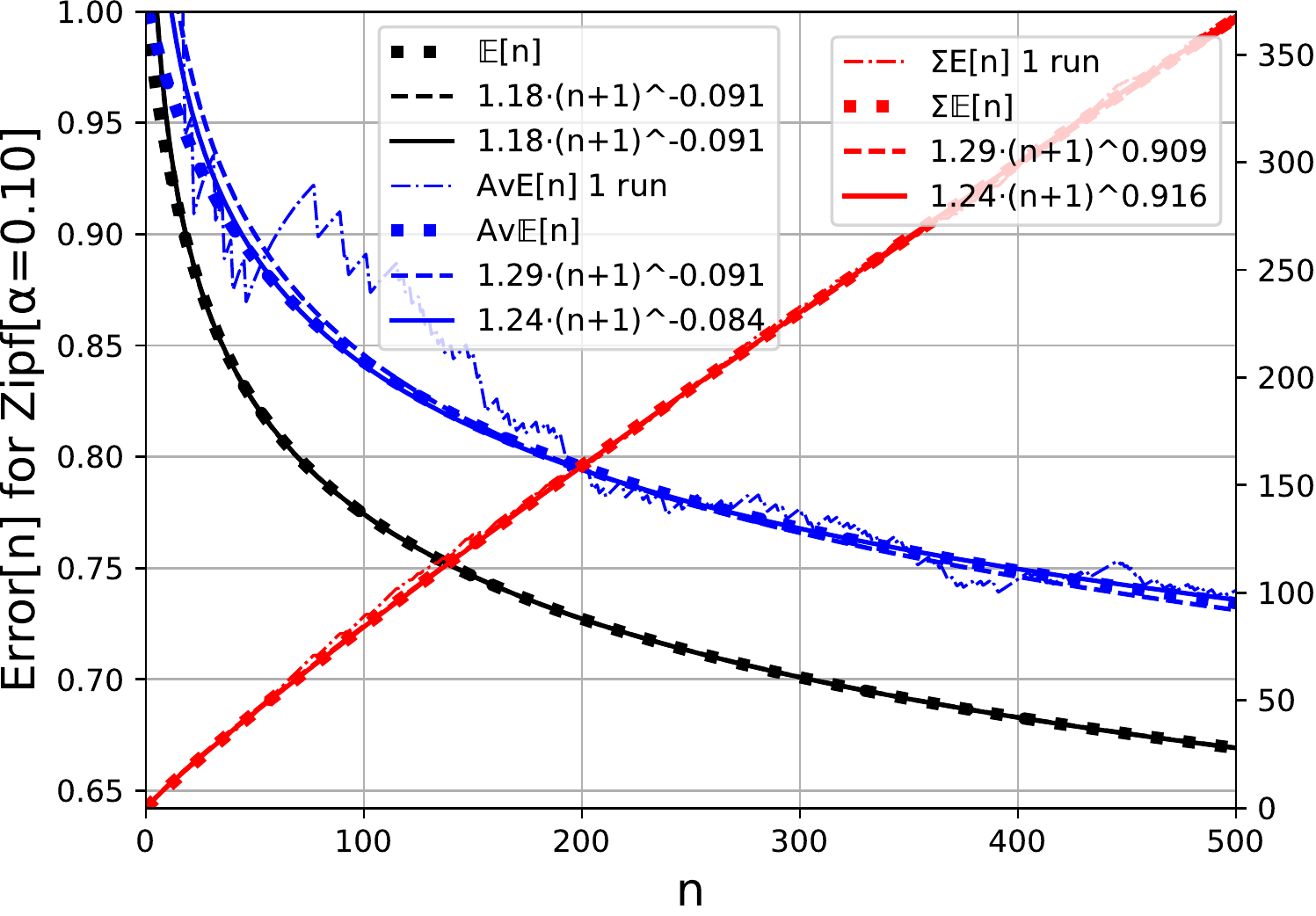}\vspace*{-2ex}
\caption{{(\bf Power Law fit to Zipf-Distributed Data)} 
for Zipf-distributed data $\P[i_n=i]=θ_i\propto i^{-(α+1)}$ 
for $α=1$ \emph{(left)} and $α=0.1$ \emph{(right)}; 
averaged over infinitely many runs (dots),
for fitted $\hat{β}$ (solid) and
theoretical $β=\fr{α}{1+α}$ (dash), 
and empirical error for a single run ($k=1$, dashdot).
}\label{fig:PowZipf}\vspace*{-2ex}
\end{center}
\end{figure*}

\begin{figure*}[ptbh!]
\begin{center}
\includegraphics[width=0.49\textwidth]{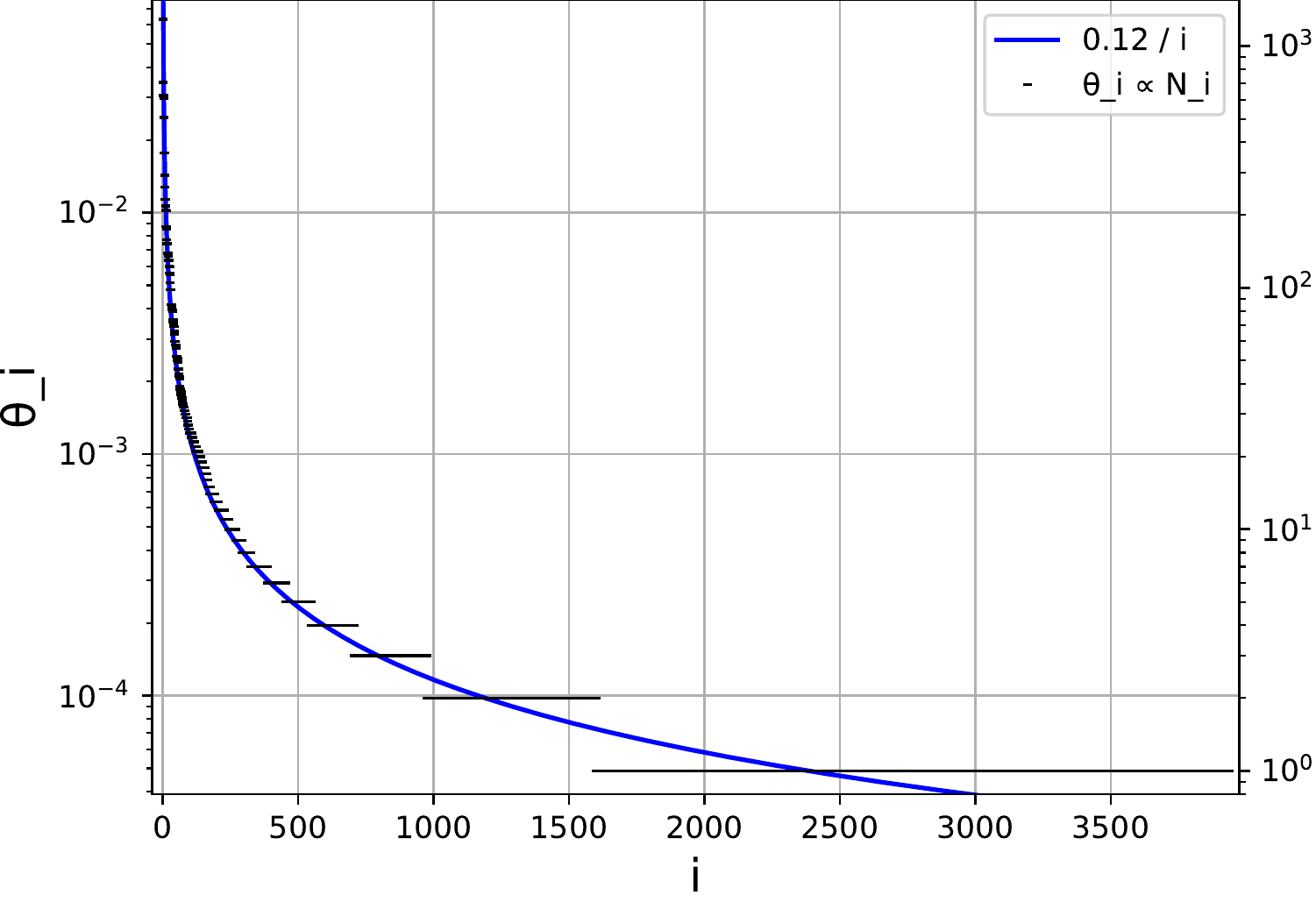}~~%
\includegraphics[width=0.49\textwidth]{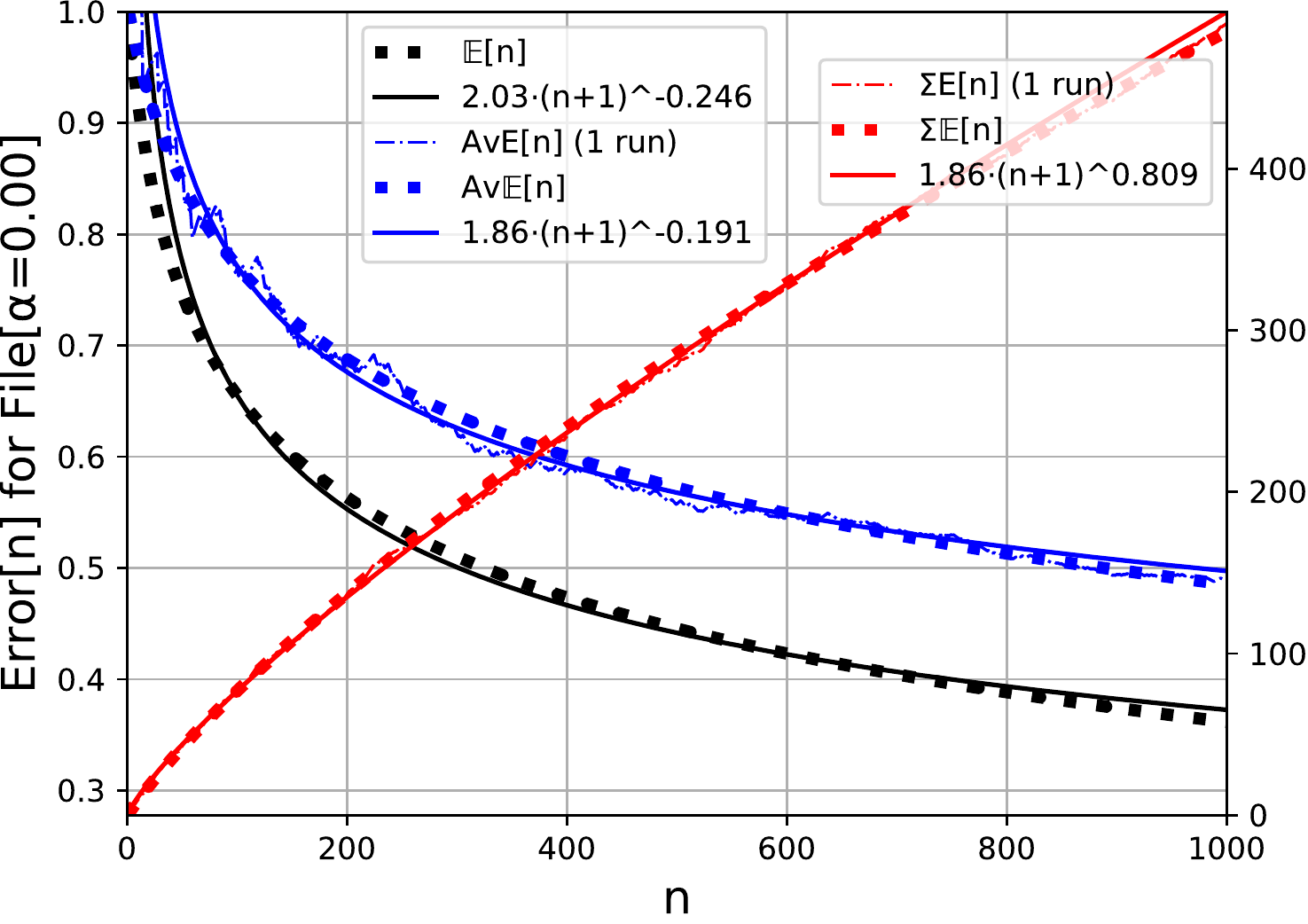}\vspace*{-2ex}
\caption{{(\bf Word-Frequency in Text File, Learning Curve, Power Law)} \emph{(left)} 
Log-linear plot of the relative (left scale) and absolute (right scale) frequency of 
words in the first 20469 words in file `book1' of the Calgary Corpus, and fitted Zipf law.
\emph{(right)} Power law fit to learning curve for this data set for a word classification task. 
}\label{fig:TextZipf}\vspace*{-2ex}
\end{center}
\end{figure*}

We performed some control experiments to verify the correctness of the theory and claims, 
and the accuracy of the theoretical expressions.

\paradot{Uniform and Zipf-distributed data} 
Figures~\ref{fig:unif} and\ref{fig:unifA} plot learning curves for 
uniformly distributed data $\P[i_n=i]=θ_i=\fr1m$ for $i≤m=10$ averaged over $k=1,10,100,1000$ runs. 
Figure~\ref{fig:zipf} and \ref{fig:zipfA} plot learning curves for 
Zipf-distributed data $\P[i_n=i]=θ_i\propto i^{-(α+1)}$ for $α=1$, also averaged over $k=1,10,100,1000$ runs. 
In both cases, 1000 synthetic data sets $\cD_{50}$ have been generated, and the average is taken over $k$ of them.
Various errors are plotted as functions of the sample index/size $n$.
The crosses are the instantaneous errors $\IE_n$ averaged over $k$ runs.
The black curves are the exact expected instantaneous error $\EE_n$.
The shaded regions are 1 standard deviation $σ_n/\sqrt{k}$ from the theory (not empirical).
Similarly the blue dots, lines, and shadings are the 
time-averaged errors $\AE_N$, 
their exact expectation $\E[\AE_N]$ and 
theoretical standard error $\sqrt{\Var[\smash{\AE_N}]/k}$.
The red triangles, lines, and shadings are the empirical cumulative errors $∑_{n=0}^N\IE[n]≡N\AE_N$, their exact expectation $∑_{n=0}^N\EE[n]≡N\E[\AE_N]$, and theoretical standard error.
The dashed lines connecting the empirical errors are for better visibility (only).

\paradot{Fitting power laws to learning curves} 
We now fit power laws to the learning curves of (exactly) Zipf distributed data.
Figure~\ref{fig:PowZipf} shows fits for synthetic data with Zipf-exponents $α=1$ and $α=0.1$.
The fit is ``perfect'' except for very small values of $n$.
This is consistent with our approximation, which is good for $nθ_1\gg 1$.
Theoretically we expect and empirically we found the approximation \eqref{eq:errapprox} to be good for $nθ_1\gtrsim 2$.
For $α=1$ we have $θ_1=0.5$, hence the approximation should be good for $n\gtrsim 4$,
while for $α=0.1$ we have $θ_1~\dot=~0.067$, hence the approximation should be good for $n\gtrsim 30$;
both are consistent with the plots. To avoid clutter we only present expected curves.
They perfectly match the averaged curves over infinitely many runs anyway (see Figures~\ref{fig:zipf}\&\ref{fig:zipfA}).
The fitted power law exponents $\hat{β}$ are also close to 
the theoretical predictions $β=\fr{α}{1+α}$ ($\fr12=0.5$ for $α=1$ and $\fr1{11}~\dot=~0.091$ for $α=0.1$).

\paradot{Text data} 
It is well-known that the frequency of a word in typical texts is about inversely proportional to its rank in the frequency table:
The most frequent word (`the') occurs about twice as often as the second most frequent word (`a'), about three times as often as the third most frequent word, and so on.
That is, word frequency follows a Zipf distribution with $α≈0$.
Figures~\ref{fig:TextZipf}\&\ref{fig:TextZipfA} (left) show the frequency distributions of the first 20469 words in file `book1' of the Calgary Corpus.
Apart from the steps, caused by word frequencies being integers, the distribution is very close to Zipf.
Note that more than half of the words only appear once.
Figures~\ref{fig:TextZipf} (right) shows the learning curves for any word classification task.
The power-law fit is reasonably good, but not perfect.
The reason is the step structure of especially rare words.
Indeed, many $θ$s are equal, and only finitely many are non-zero, 
so the learning curve is a finite superposition of exponentials as in \eqref{eq:seerr}.
For moderate $n$ this mixture amalgamates to an approximately power law.
For large $n$, the error decays exponentially as $\exp(-θ_{min}n)$.
Indeed, for larger $n$, Figures~\ref{fig:TextZipfA} shows that the power fit becomes worse,
and the true error decays faster than the fit power law.
Note that for $α≤0$, $i^{-(α+1)}$ is not summable, 
hence \emph{any} such distribution must break down after some $i$,
our approximation becomes invalid, 
and $β≤0$ makes no sense in any case.

\section{Extensions}\label{sec:Extend}

In the following we discuss some potential extensions of the toy model.
Some look feasible, others are hard or wishful thinking.
We discuss the more realistic case of noisy labels, other loss functions, 
continuous features, and more realistic models that generalize, 
e.g.\ deep learning algorithms.

\paradot{Noisy labels} 
In most machine learning applications, labels (or more general targets) are themselves noisy, 
not just the feature vector $x∈\cX$, e.g.\ $y=h(x)+${\it Noise}. The major implications are as follows:
\begin{itemize}\parskip=0ex\parsep=0ex\itemsep=0ex
\item[(a)] The learning algorithms need to be a bit smarter than just memorization, \\
           e.g.\ predicting the average or by majority. 
\item[(b)] Due to the label noise, the error cannot converge to $0$ anymore but to the intrinsic ``entropy'', 
which should be subtracted before studying scaling.
\item[(c)] For absolute (locally quadratic) loss there will be an extra $n^{-1/2}$ ($n^{-1}$) additive error term due to parameter estimation error,
hence 
\item[(d)] the instantaneous loss will not decay exponentially anymore even if the model is finite.
\item[(e)] Otherwise the scaling laws for Zipf data are \emph{unchanged}.
\end{itemize}
In summary, the error/loss should be a sum of 3 terms, at least conceptually:
\begin{itemize}\parskip=0ex\parsep=0ex\itemsep=0ex
\item[(1)] The inherent entropy in the data, 
\item[(2)] the parameter learning rate $n^{-1/2}$ for absolute loss, squared, i.e.\ $n^{-1}$ for (locally) quadratic loss, 
\item[(3)] the same power law $n^{-β}$ as in the deterministic case.
\end{itemize}
In Appendix~\ref{app:Noise} we verify claims (b,c,d,e) for our toy model extended to noisy binary classification for square loss.
What is remarkable is that the \emph{instantaneous} square loss $\Loss_n(A)$ for noisy labels  
turns out to include a term proportional to the \emph{time-averaged} (0-1) error $\E[\AE_n]=$\eqref{eq:eae} in the deterministic case.
But this ``magically'' ensures (c,d,e), 
since $\E[\AE_n]\eqam\max\{\EE_n,\fr1n\}$, 
at least for the choices of $\vth$ discussed in Section~\ref{sec:Rates}.
For instance, for a finite model, $\Loss_n(A)\eqam\E[\AE_n]\eqam\fr1n$.

\paradot{Other loss functions} 
For our deterministic toy model, the loss function has little to no influence on the results as discussed in Appendix~\ref{app:Loss}.
For noisy labels, this also seems to be the case,
except that $n^{-β_0}$ is now the fastest possible decay, 
with $β_0$ \emph{depending} on the loss-function: $β_0=\fr12$ for absolute loss and $β_0=1$ for locally quadratic loss such as KL and square. 
Loss functions with any (other) value of $β_0>0$ are possible but rare.
This is another potential universality of scaling laws, 
their independence from the loss function for large models.

\paradot{Continuous features} 
Countable feature spaces have some applications, e.g.\ in NLP, 
words can be identified with integer features $i∈ℕ$.
In most applications, feature spaces are (effectively) continuous, 
often vector spaces $ℝ^d$, and no feature ever repeats exactly ($x_n≠x_m$ for $n≠m$).
A simple model with a continuous domain is the Dirichlet Process, 
or the essentially equivalent Chinese Restaurant Process (CRP) and Stick-Breaking process.
In the CRP, the continuous domain is essentially reduced to an 
exponentially distributed countable number of sticks=features,
leading to power law learning curves $n^{-β}$,
but the exponent is restricted to $β=1$, which is too limiting.
But even the CRP is not exactly a special case of our toy model and much harder to analyse.
In some form of ``mean-field'' approximation it reduces to a special case of our model.
The generalized 2-parameter Poison Dirichlet Process \cite{Hutter:10pdpx} also only leads to $β=1$.
Finding analytically tractable models with continuous features that exhibit interesting
learning curves remains an open problem.

\paradot{Generalizing algorithms} 
Proper models/algorithms for continuous features need to generalize from observed inputs to similar future not-yet-observed inputs, 
which is at the heart of virtually all interesting machine learning model/algorithms.
Such models are much more varied and harder to analyze. 
If the domain could be partitioned into countably many cells, 
each cell containing only sufficiently similar features,
and this can be done a-priori and is fixed independent of the actually realized data $\cD_n$ 
and most importantly independent of the data size $n$,
we arrive back at our countable toy model (usually with noisy labels) and our analysis (nearly) applies.
But it is more plausible that a suitable partitioning, e.g.\ clustering of data, 
is in itself data (size) dependent, and hence will affect the scaling.
A more interesting non-parametric model, potentially amenable to theoretical analysis, is $k$-Nearest-Neighbors (kNN),
likely with interesting learning curves.
The `perfect prediction for exact repetition' in our toy model can be viewed as an abstraction 
of `classify features in the same cell alike' which itself is a toy model for 
`classify similar observations alike or similarly',
so maybe some of our findings or analysis tools approximately transfer.

\paradot{Deep learning} 
(Deep) neural networks are a particularly powerful class of models/algorithms that can generalize, 
but are also notoriously difficult to theoretically analyse.
It may be a long way from our toy model to a similar analysis of NNs.
Furthermore we have not at all considered the equally interesting questions of scaling with model size and compute.

\section{Discussion}\label{sec:Disc}

\paradot{Summary}
We introduced a very simple model that can exhibit power laws (decrease of error with data size)
consistent with recent findings in deep learning.
The model is plausibly the simplest such model, 
and that choice was deliberate to not get bogged down with intractable math 
or forced into crude approximations or bounds at this early stage of investigation.
Many but not all data distributions lead to power laws.
We do not know whether the discovered specific relation between the Zipf exponent $α$ 
and the power law exponent $β=α/(1+α)$ is an artifact of the model, 
or has wider validity beyond this model.
The signal-to-noise ratio for the time-averaged error tend to zero,
which implies that a single experimental run suffices for stable results.

\paradot{Limitations}
The toy model studied in this work is admittedly totally unrealistic as a Deep Learning model, 
but we believe it captures the (or at least a) true reason for the observed scaling laws w.r.t.\ data.
Whether it has any predictive power, or can be generalized to NNs and/or 
scaling laws for model size and/or compute, is beyond the scope of this paper. 
We hope that this initial investigation spurs more advanced theoretical investigations,
and ultimately lead to predictive models.
We have outlined some ideas in Section~\ref{sec:Extend}, some (more) are hopefully feasible.
In any case, finding the simplest model which captures the essence is a necessary first step, 
and we believe our toy model fits this bill.

\paradot{Applications}
Besides providing scientific insight, 
a good theoretical understanding of scaling laws could ultimately help tune network and algorithm parameters in a more principled way,
and thus save significant compute for finding good large NNs by reducing hyper-parameter sweeps.
The cost of training recent models has reached millions of dollars and can exhaust and exceed even FAANGs computational resources.

\paradot{Acknowledgements}
I thank David Budden and Jörg Bornschein for encouraging me to look into the topic of scaling laws,
and for interesting discussions.

\section*{References}\label{sec:Bib}
\addcontentsline{toc}{section}{\refname}
\def\refname{\vspace{-4ex}}
\bibliographystyle{alpha} 
\begin{small}
\newcommand{\etalchar}[1]{$^{#1}$}

\end{small}

\appendix

\section{Other Loss Functions}\label{app:Loss}

We can (slightly) generalize the learning algorithm $A$ to other loss functions and behaviors on $i_{n+1}\not\in i_{1:n}$.
We continue to assume that $A'$ suffers $\Loss_n=0$ if $i∈i_{1:n}$ by using stored $\cD_n$.
Assume $A'$ suffers $\Loss_n$ if $i_{n+1}\not\in i_{1:n}$, then
\begin{align*} 
  \E[\Loss_n] ~&=~ \E[\Loss_n|i_{n+1}\not\in i_{1:n}]⋅\P[i_{n+1}\not\in i_{1:n}] + 0⋅\P[i∈i_{1:n}] ~=~ \E[\Loss_n|i_{n+1}\not\in i_{1:n}]⋅\EE_n
\end{align*}
The second factor is our primary object of study. The first factor is often constant or bounded by constants:
For instance for $A$ in \eqref{eq:alg}, $\Loss_n=1$ if $i_{n+1}\not\in i_{1:n}$, hence $\E[\Loss_n]=\EE_n$.
Assume a classification problem with $K$ labels $\cY=\{0,...,K-1\}$,
and modify $A$ to randomize its output (samples $y∈\cY$ uniformly) if $i_{n+1}\not\in i_{1:n}$,
then there is a $1/K$ chance to accidentally predict the correct label,
hence $\E[\Loss_n|i_{n+1}\not\in i_{1:n}]=1-1/K$, where the expectation is now also w.r.t.\ the algorithm's randomness.
For continuous $\cY$ and if $A'$ samples $y$ from any (non-atomic) density over $\cY$ if $i_{n+1}\not\in i_{1:n}$, 
the probability of accidentally correctly predicting $y_{n+1}$ is $0$, hence $\E[\Loss_n]=\EE_n$.
For binary classification ($K=2$) we could also let $A'$ predict $y=\fr12$ and use absolute loss $\Loss_n=|y-y_{n+1}|$,
in which case $\E[\Loss_n|i_{n+1}\not\in i_{1:n}]=\fr12$.
If $A'$ instead samples $y$ uniformly from $[0;1]$,
then $\E[\Loss_n|i_{n+1}\not\in i_{1:n}]=∫_0^1|y-y_{n+1}|dy=\fr14$. 
If $\cY=[0;1]$, then $\E[\Loss_n|i_{n+1}\not\in i_{1:n}]=∫_0^1|y-y_{n+1}|dy=\fr14+(\fr12-y_{n+1})^2$, 
hence $\fr14\EE_n≤\E[\Loss_n]≤\fr12\EE_n$.

More generally, for any compact and uniformly rounded set $\cY⊆ℝ^d$, 
for any loss of the form $\Loss_n:=\ell(||y-y_{n+1}||)$, 
for any norm $||⋅||$,
for any continuous strictly increasing $\ell≥0$,
and $A'$ sampling $y$ from any density $p_{alg}(y)>0$ over $\cY$ if $i_{n+1}\not\in i_{1:n}$, 
we have, for some constants $c_1,c_2>0$,
\begin{align*}
  & c_1\EE_n ~≤~ \E[\Loss_n] ~≤~ c_2\EE_n, ~~~&\text{or}~~~ \E[\Loss_n] ~\eqm~ \EE_n ~~\text{for short}
\end{align*}
and this fact holds even more generally.
Since a multiplicative constant in the loss is irrelevant from a scaling perspective, 
all scaling results for $\EE_n$ also apply to this (slightly) more general setting.

The proof of this is as follows: A uniformly rounded set by definition can be represented as a union of 
$ε$-balls for some fixed $ε>0$, i.e.\ $\cY=\bigcup_{\tilde y∈\tilde\cY} B_ε(\tilde y)$ for some $\tilde\cY$,
where $B_ε(\tilde y):=\{y:||y-\tilde y||≤ε\}$. 
Then
\begin{align}
  \E[\Loss_n|i_{n+1}\not\in i_{1:n}] ~&=~ ∫_\cY\ell(||y-y_{n+1}||)p_{alg}(y)dy \label{eq:gloss}\\
  ~&\stackrel{(a)}{≥}~ δ∫_{B_ε(\tilde y)}\ell(||y-y_{n+1}||)dy \nonumber\\
  ~&\stackrel{(b)}{≥}~ δ∫_{B_ε(y_{n+1})}\ell(||y-y_{n+1}||)dy \nonumber\\
  ~&\stackrel{(c)}{=}~ δ∫_{B_ε(0)}\ell(||z||)dz ~=:~ c_1 ~\stackrel{(d)}{>}~ 0 \nonumber 
\end{align}
In (a), $\tilde y∈\tilde\cY$ is chosen such that $y_{n+1}∈B_ε(\tilde y)⊆\cY$,
and $p_{alg}>δ>0$, since $p_{alg}>0$ and $\cY$ is compact.
In (b), $B_ε(y_{n+1})$ can be obtained from $B_ε(\tilde y)$ by cutting out the moon 
$B_ε(\tilde y)\setminus B_ε(y_{n+1})$, and point-mirroring the moon at $\fr12(\tilde y+y_{n+1})$.
The flip brings every point closer to $y_{n+1}$,
hence decreases the integral since $\ell$ is monotone increasing.
(c) just recenters the integral, which now is obviously independent of $y_{n+1}$.
It is non-zero (d), since $\ell$ is strictly increasing. 
Since $\ell$ is continuous and $\cY$ is compact, $\ell$ is upper bounded by $\ell_{max}<∞$.
This immediately implies \eqref{eq:gloss} is upper bounded by $c_2:=\ell_{max}<∞$.

\section{Derivation of Expectation and Variance}\label{app:Var}

\paradot{Expectation}
Recall the error of (the basic form of) Algorithm $A$ is $\IE_n=[\![i_{n+1}\not\in i_{1:n}]\!]$.
Hence the probability that Algorithm $A$ makes an error under distribution $\vth$ given data $\cD_n$ is 
\begin{align}\label{eq:err}
  \E[\IE_n|\cD_n] ~=~ \P[A(i_{n+1},\cD_n)≠y_{n+1}|\cD_n] 
  ~=~ ∑_{i\not\in i_{1:n}}\P[i_{n+1}=i]
  ~=~ ∑_{i\not\in i_{1:n}}θ_i
\end{align}
The expectaion of this w.r.t.\ $\cD_n$ is 
\begin{align*}
  \EE_n ~&:=~ \E[\IE_n] ~=~ \E[\E[\IE_n|\cD_n]] 
             ~=~ ∑_{i_{1:n}}\P[A(i_{n+1},\cD_n)≠y_{n+1}|\cD_n]\P[\cD_n] \\
             ~&=~ ∑_{i_{1:n}} \Big(∑_{i\not\in i_{1:n}}θ_i\Big)∏_{t=1}^n θ_{i_t}
             ~=~ ∑_{i_{1:n}} \Big(∑_{i=1}^∞[\![i≠i_1∧...∧i≠i_n]\!]θ_i\Big)∏_{t=1}^n θ_{i_t} \nonumber\\
             ~&=~ ∑_{i=1}^∞ θ_i ∑_{i_{1:n}} ∏_{t=1}^n [\![i≠i_t]\!]θ_{i_t}
             ~=~ ∑_{i=1}^∞ θ_i ∏_{t=1}^n ∑_{i_t≠i} θ_{i_t}
             ~=~ ∑_{i=1}^∞ θ_i (1-θ_i)^n \nonumber
\end{align*}
The result can actually more easily be derived as
\begin{align}
  \EE_n ~&=~ \P[i_{n+1}\not\in i_{1:n}]
             ~=~ ∑_{i=1}^∞ \P[i_{n+1}=i∧i_1≠i∧...∧i_n≠i] \nonumber\\
             ~&=~ ∑_{i=1}^∞ \P[i_{n+1}=i]∏_{t=1}^n \P[i_t≠i]
             ~=~ ∑_{i=1}^∞ θ_i (1-θ_i)^n \label{eq:eerrA}
\end{align}
but the former derivation is more suitable for generalization to other loss functions and noisy labels.

\paradot{Approximation}
Let $f:(0;∞)→(0;∞)$ be a continuously differentiable and decreasing extension of $\vth:ℕ→ℝ$,
i.e.\ $f(i):=θ_i$ and $f'(x)<0$. Let $g(x):=f(x)e^{-nf(x)}$. Since $u\mapsto ue^{-nu}$ is unimodal
with maximum $1/en$ and at $u=1/n$ and $f$ is monotone, 
$g(x)$ is unimodal with maximum $g_{max}=1/en$ at $x_{max}=f^{-1}(\fr1n)$.
We hence can use \eqref{eq:incdec} (any $a∈[0;1]$) to upper bound the sum in \eqref{eq:eerrA} by an integral as follows:
\begin{align*}
  \EE_n ~&\stackrel{\eqref{eq:eerrA}}=~ ∑_{i=1}^∞ θ_i (1-θ_i)^n
  ~≤~  ∑_{i=1}^∞ θ_i e^{-nθ_i}
  ~=~ ∑_{i=1}^∞ f(i)e^{-nf(i)} \nonumber\\
  ~&\stackrel{\eqref{eq:incdec}}{≤}~ g_{max} + \int_a^∞ f(x)e^{-nf(x)}dx
  ~=~ \frac1{en} + \int_0^{f(a)} {u e^{-nu}du\over|f'(f^{-1}(u))|}
\end{align*}
where the last equality follows from a reparametrization $u=f(x)$ and $f(∞)=0$ and $dx=du/f'(x)$ and $f'<0$.

For a lower bound, we need a lower bound on $(1-θ_i)^n$.
For $0≤x≤2ε$ we have 
\begin{align*}
  e^{-x} ~≤~ 1-x+\fr12 x^2 ~=~ 1-(1-\fr12 x)x ~≤~ 1-(1\!-\!ε)x
\end{align*}
Inserting $x=θ_i/(1-ε)$, we get 
\begin{align*}
  1-θ_i ~≥~ e^{-θ_i/(1-ε)} ~~~\text{for}~~~ θ_i≤2ε(1-ε)
\end{align*}
Let $i_0$ be an index such that $θ_{i_0}≤2ε(1-ε)$.
We define $\tilde n:=n/(1-ε)$. 
That $\tilde n$ is not an integer is no problem,
and could even be avoided by renormalizing $θ_i$ instead.
Similarly as for the upper bound, we get a lower bound
\begin{align*}
  \EE_n ~&≥~ ∑_{i=i_0}^∞ θ_i (1-θ_i)^n
  ~≥~  ∑_{i=i_0}^∞ θ_i e^{-\tilde n θ_i}
  ~=~  -∑_{i=1}^{i_0-1} θ_i e^{-\tilde n θ_i} + ∑_{i=1}^∞ θ_i e^{-\tilde n θ_i} \\
  ~&≥~ - ∑_{i=1}^{i_0-1} θ_i e^{-\tilde n θ_{i_0}} - \tilde g_{max} + \int_a^∞ f(x)e^{-\tilde n f(x)}dx \\
  ~&≥~ - e^{-\tilde n θ_{i_0}} - \frac1{e\tilde n} + \int_0^{f(a)} {u e^{-\tilde n u}du\over|f'(f^{-1}(u))|}
\end{align*}
Let us choose $i_0$ such that $θ_{i_0}≥ε(1-ε)$, which is possible as long as $θ_i≥\fr12 θ_{i-1}$.
This finally leads to 
\begin{align*}
  \EE_n ~≥~ - e^{-εn} - \frac1{en} + \int_0^{f(a)} {u e^{-\tilde n u}du\over|f'(f^{-1}(u))|}
\end{align*}
Since we can choose $ε$ arbitrarily small,
combining both bounds, and choosing $a→0$ and $f$ such that $f(x)→∞$ for $x→0$, we have 
\begin{align}\label{eq:eerrapproxA}
  \Big|\EE_n - \EE_n^{∫}\Big| ~≤~  \frac1{en} + o(1/n) ~~~\text{with}~~~ 
  \EE_n^{∫} ~:=~ \int_0^∞ {u e^{-nu}du\over|f'(f^{-1}(u))|}
\end{align}
The integral is dominated by $u\lesssim\fr1n$, so for large $n$ is determined by the asymptotics
of $f'(f^{-1}(u))$ for $u→0$.
Assume
\begin{align}\label{eq:ffapproxA}
  |f'(f^{-1}(u))| ~≈~ c'u^δ ~~~\text{for ~~$u→0$~~ for some ~~$c'$ and $δ$}
\end{align}
Substitution $u=v/n$ leads to
\begin{align}\label{eq:eerrintA}
  \EE_n^{∫} ~≈~ \int_0^∞ {u e^{-n u}\over c'u^δ}du ~=~ {n^{δ-2}\over c'}\int_0^∞ v^{1-δ} e^{-v}dv ~=~ \fr{Γ(2-δ)}{c'}n^{δ-2}
\end{align}
where $Γ$ is the Gamma function.

\paradot{Zipf distribution}
For Zipf-distributed $θ_i=αi^{-(α+1)}$, let $u=f(x)=α⋅x^{-(α+1)}$. 
This implies 
\begin{align*}
  x = f^{-1}(u)=(u/α)^{-\fr1{1+α}} ~~~\text{and}~~~ f'(x) = -α(α+1)x^{-(α+2)} = -α(α+1)(u/α)^{\fr{α+2}{α+1}}
\end{align*}
hence 
approximation \eqref{eq:ffapproxA} is actually exact with $δ=\fr{α+2}{α+1}$ and $c'=α(α+1)/α^{(α+2)/(α+1)}$ leading to
\begin{align*}
  \EE_n^{∫} = c_α n^{-β} ~~~\text{with}~~~
   c_α = \fr{Γ(2-δ)}{c'}  = \fr{α^{1/(1+α)}}{1+α}Γ(\fr{α}{1+α})
   ~~~\text{and}~~~ β = δ-2 = \fr{α}{1+α}
\end{align*}
Note that $\tilde n^{-β}=n^{-β}+o(\fr1n)$ for e.g.\ $ε=\fr2n\ln n$, and $e^{-εn}=1/n^2$.
Numerically one can check that $c_α≤1.214$ for all $α>0$ and $c_α≥0.886$ for (the interesting) $α≤1$,
that is, $c_α$ is nearly independent of $α$. 
$c_1=\fr12\sqrt\pi~\dot=~0.886$ and $c_{0.1}=1.177$ are in excellent agreement with the fit curves in Figure~\ref{fig:PowZipf}.

\paradot{Time-averaged expectation and variance}
We now consider the time-averaged error
\begin{align*}
  \AE_N ~:&=~ {1\over N}\sum_{n=0}^{N-1}\IE_n
\end{align*}
We derive the expressions for its expectation $\E[\AE_N]$ and variance $\Var[\AE_N]$ stated in Section~\ref{sec:Var}.
The expectation is trivial:
\begin{align*}
  \E[\AE_N] ~&=~ {1\over N}\sum_{n=0}^{N-1}\E[\IE_n]
  ~=~ {1\over N}\sum_{i=1}^∞ θ_i\sum_{n=0}^{N-1}(1-θ_i)^n 
  ~=~ {1\over N}\sum_{i=1}^∞ [1-(1-θ_i)^N] 
\end{align*}
For the variance, we first compute $\AE_N^2$, then $\E[\AE_N^2]$, then $\Var[\AE_N]=\E[\AE_N^2]-\E[\AE_N]^2$:
\begin{align*}
  \AE_N^2 ~&=~ {1\over N^2}\sum_{n=0}^{N-1}\sum_{m=0}^{N-1}\IE_n \IE_m
  ~\stackrel{(b)}=~ {2\over N^2}\sum_{n=0}^{N-1}\sum_{m=0}^{n-1}\IE_n \IE_m + {1\over N^2}\sum_{n=0}^{N-1}\IE_n^2 \\ 
\end{align*}
where $(b)$ breaks up the double sum into lower=upper and diagonal terms.
Since $m<n$ we have
\begin{align*}
  \IE_n⋅\IE_m ~&=~ [\![i_{n+1}\not\in i_{1:n}]\!]⋅[\![i_{m+1}\not\in i_{1:m}]\!] \\
  ~&=~ \sum_{i=1}^∞ [\![i_{n+1}=i∧i_n≠i∧...∧i_1≠i]\!] ⋅ \sum_{j=1}^∞ [\![i_{m+1}=j∧i_m≠j∧...∧i_1≠j]\!] \\
  ~&=~ \sum_{i,j} [\![i_{n+1}=i∧i_n≠i∧...∧i_{m+2}≠i∧i_{m+1}{=j\atop ≠i}∧i_m{≠j\atop ≠i}∧...∧i_1{≠j\atop ≠i}]\!]
\end{align*}
For $i=j$, $i_{m+1}{=j\atop ≠i}$ (meaning $i_{m+1}=j∧i_{m+1}≠i$) is a contradiction, so we can limit the sum to $i≠j$.
Taking the expectation and noting that $\IE_n∈\{0,1\}$ and all $i_t$ are independent and $\P[i_t=i]=θ_i$ we get
\begin{align*}
  \E[\IE_n⋅\IE_m] ~&=~ \sum_{i≠j} \P[i_{n+1}=i]\P[i_n≠i]...\P[i_{m+2}≠i]\P[i_{m+1}{=j\atop ≠i}]\P[i_m{≠j\atop ≠i}]...\P[i_1{≠j\atop ≠i}] \\
  ~&=~ \sum_{i≠j} θ_i(1-θ_i)^{n-m-1}θ_j(1-θ_i-θ_j)^m,~~~\text{hence}
\end{align*}\vspace{-2ex}
\begin{align*}
  \E[\sum_{n=0}^{N-1}\sum_{m=0}^{n-1}\IE_n \IE_m] ~&=~ \sum_{i≠j} b_{ijN} = \fr12\sum_{i≠j} b_{ijN}+b_{jiN},~~~\text{where}
\end{align*}\vspace{-2ex}
\begin{align*}
  b_{ijN} ~&:=~ \sum_{n=0}^{N-1}\sum_{m=0}^{n-1} θ_i(1-θ_i)^{n-m-1}θ_j(1-θ_i-θ_j)^m \\
  ~&=~ θ_i θ_j\sum_{m=0}^{N-1}(1-θ_i-θ_j)^m\sum_{n=m+1}^{N-1}(1-θ_i)^{n-m-1} \\
  ~&=~ θ_i θ_j\sum_{m=0}^{N-1}(1-θ_i-θ_j)^m{1\over θ_i}[1-(1-θ_i)^{N-m-1}] \\
  ~&=~ θ_j\sum_{m=0}^{N-1}(1-θ_i-θ_j)^m ~-~ θ_j(1-θ_i)^{N-1}\sum_{m=0}^{N-1}\Big({1-θ_i-θ_j\over 1-θ_i}\Big)^m \\
  ~&=~ θ_j{1\over θ_i+θ_j}[1-(1-θ_i-θ_j)^N] ~-~ θ_j(1-θ_i)^{N-1}{1-θ_i\over θ_j}\bigg[1-\Big({1-θ_i-θ_j\over 1-θ_i}\Big)^N\bigg] \\
  ~&=~ {θ_j\over θ_i+θ_j}[1-(1-θ_i-θ_j)^N] ~-~ (1-θ_i)^N ~+~ (1-θ_i-θ_j)^N,~~~\text{hence} \\
  b_{ijN} &+ b_{jiN} ~=~ 1 - (1-θ_i)^N - (1-θ_j)^N + (1-θ_i-θ_j)^N
\end{align*}
The diagonal term is easy:
\begin{align*}
  {1\over N^2}\sum_{n=0}^{N-1}\E[\IE_n^2] 
  ~&=~ {1\over N^2}\sum_{n=0}^{N-1}\E[\IE_n] 
  ~=~ {1\over N}\sum_{n=0}^{N-1}\E[\AE_N] ~=~ {1\over N^2}\sum_{i=1}^∞ 1-(1-θ_i)^N
\end{align*}
Putting everything together, non-diagonal $i≠j$ and diagonal $i=j$ expressions we get our final expression
\begin{align*}
  \E[\AE_N^2] ~&=~ {1\over N^2}\sum_{i≠j}[1-(1-θ_i)^N-(1-θ_j)^N+(1-θ_i-θ_j)^N] ~+~ {1\over N^2}\sum_{i=1}^∞[1-(1-θ_i)^N]
\end{align*}
In order to get the variance of $\AE_N$ we have to subtract the squared expected error
\begin{align*}
  \E[\AE_N]^2 ~&=~ \Big({1\over N}\sum_{i=1}^∞ 1-(1-θ_i)^N\Big)^2 ~=~ {1\over N^2}\sum_{i,j}[1-(1-θ_i)^N]⋅[1-(1-θ_j)^N] \\
  ~&=~ {1\over N^2}\sum_{i≠j}[1-(1-θ_i)^N-(1-θ_j)^N+(1-θ_i)^N(1-θ_j)^N] \\
   &   ~+~ {1\over N^2}\sum_{i=1}^∞[1-2(1-θ_i)^N+(1-θ_i)^{2N}]
\end{align*}
where we expanded the product and separated the $i≠j$ from the $i=j$ terms, which now easily leads to 
\begin{align*}
  \Var[\AE_N] ~&=~ \E[\AE_N^2] - \E[\AE_N]^2 \\
  ~&=~ {1\over N^2}\sum_{i≠j}[(1-θ_i-θ_j)^N-(1-θ_i)^N(1-θ_j)^N] ~+~ {1\over N^2}\sum_{i=1}^∞[(1-θ_i)^N-(1-θ_i)^{2N}]
\end{align*}

\paradot{Approximation}
We can approximate the variance similarly to the expectation $\EE_n$.
We only provide a heuristic derivation analogous to \eqref{eq:errapprox}:
\begin{align*} 
  \Var[\AE_N] ~&\stackrel{(a)}{≈}~ {1\over N^2}\sum_{i≠j}[e^{-(θ_i-θ_j)N}-e^{-θ_i N}e^{-θ_j N}] ~+~ {1\over N^2}\sum_{i=1}^∞[e^{-θ_i N}-e^{-2θ_i N}] \\
              ~&\stackrel{(b)}{≈}~ 0 ~+~ {1\over N^2}\int_1^∞ e^{-f(x)N}-e^{-2f(x)N}dx 
               ~\stackrel{(c)}=~ {1\over N^2}\int_0^{θ_1} {e^{-uN}-e^{-2uN}\over|f'(f^{-1}(u))|}du \\
              ~&\stackrel{(d)×}{≈}~ {1\over N^2|f'(f^{-1}(\fr{\ln2}N))|}\int_0^{θ_1} e^{-uN}-e^{-2uN}du 
               ~\stackrel{(e)}{≈}~ {1\over 2N^3|f'(f^{-1}(\fr{\ln2}N))|} \\
              ~&\eqam~ {1\over 2N}\EE_N ~\left\{ {\eqam~~~ \fr1N\E[\AE_N] ~~\text{if}~~ \EE_N\eqam N^{-β}
                                                              \atop ≤~ \fr1{2N}\E[\AE_N] ~~\text{always}~~~~~~~~} \right.
\end{align*}
$(a)$ follows from $1-θ_i≈e^{-θ_i}$, 
$(b)$ by setting $f(i):=θ_i$, and replacing the sums by integrals,
$(c)$ follows from a reparametrization $u=f(x)$ and $f(1)=θ_1$ and $f(∞)=0$ and $dx=du/f'(x)$ and $f'<0$.
The numerator $e^{-uN}-e^{-2uN}$ is maximal and (strongly) concentrated around $u=\ln2/n$, hence $u≈\ln2/n$ gives most of the integral's contribution.
Therefore in $(d)$ replacing $u$ by $\ln2/n$ in the denominator can be a reasonable approximation.
$(e)$ follows from $∫_0^{θ_1}...≈∫_0^∞...=1/2N$ for $θ_1N\gg 1$.
We could use this approximation for various concrete $f$, 
but if we substitute $u=1/N$ instead of $\ln2/N$ in $(d)$ we only make a multiplicative error $(×)$,
and the expression nicely reduces to the \emph{instantaneous} expected error $\EE_N$.
For slowly decreasing error $\EE_N\eqam N^{-β}$, we have $\E[\IE_N]\eqam\E[\AE_N]$.
In general $\E[\IE_N]≤\E[\AE_N]$, since $\EE_n$ is monotone decreasing.

\section{Noisy Labels}\label{app:Noise}

Here we generalize our model to noisy labels. 
We first derive generic expression expressions for (somewhat) general algorithm and loss.
We then instantiate them for frequency estimation and square loss.
Finally we outline how to derive similar expressions for the absolute loss.

\paradot{General loss}
Consider a binary classification problem where labels $y_t∈\{0,1\}$ are noisy.
Let $γ_i:=\P[y_t=1|i_t=i]$ be the probability that feature $i∈ℕ$ is labelled as $1$.
The probability of observing feature $i$ itself remains $θ_i:=\P[i_t=i]$ as in the deterministic case.
Algorithm $A_i:=A(i,\cD_n)∈[0;1]$ now aims to predict $γ_i$.
The square loss if predicting $A_i$ while the true label is $y$, and its expectation w.r.t.\ $γ_i$, are 
\begin{align*}
  & \Loss_n(A|\cD_n,i_{n+1}=i,y_{n+1}=y) ~=~ (y-A_i)^2 \\
  & \Loss_n(A|\cD_n,i_{n+1}=i) ~=~ γ_i(1-A_i)^2 + (1-γ_i)(0-A_i)^2 ~=~ (γ_i-A_i)^2 + γ_i(1-γ_i)
\end{align*}
The most naive learning algorithm would predict $γ_i$ from observed frequencies:
$A_i=k_i/n_i$ if feature $i$ occurred $n_i:=\#\{t≤n:i_t=i\}$ times and has label $1$ for $k_i:=\#\{t≤n:i_t=i,y_t=1\}$ times.
Obviously $k_i/n_i→γ_i$ provided $n_i→∞$, 
hence $\Loss_n(A|\cD_n,i_{n+1}=i)$ converges to the intrinsic label ``entropy'' $γ_i(1-γ_i)$, 
rather than to $0$, which has to be subtracted for a power law analysis to make sense.
Similarly the expectation of log-loss $-y\ln A_i-(1-y)\ln(1-A_i)$ w.r.t.\ $γ_i$ leads to 
Kullback-Leibler loss $\text{KL}(γ_i||A_i)$ + Entropy $H(γ_i)$.
More generally let us assume $A(i,\cD_n)$ depends (somehow) only on $k_i$ and $n_i$ (e.g.\ Laplace rule),
and hence 
\begin{align*}
  \Loss_n(A|\cD_n,i_{n+1}=i) ~=~ \ell(γ_i,A_i) ~=~ \ell(γ_i,k_i,n_i) ~=~ \Loss_n(A|k_i,n_i,i_{n+1}=i)
\end{align*}
for some function $\ell$. We now take the expectation over $\cD_n$:
\begin{align*}
  L_i ~&:=~ \Loss_n(A|i_{n+1}=i) \\
  ~&=~ \sum_{\cD_n} \Loss_n(A|\cD_n,i_{n+1}=i)\P[\cD_n] \\
  ~&=~ \sum_{n_i=0}^n\sum_{k_i=0}^{n_i}\bigg(\sum_{\cD_n:k_i,n_i\nq\nq}\P[\cD_n]\bigg)\ell(γ_i,k_i,n_i)
\end{align*}
where $\sum_{\cD_n:k_i,n_i}$ means that the sum is restricted to $\cD_n$ for which $i$ ($(i,1)$) appears $n_i$ ($k_i$) times.
The probability of each of this happening is binomial:

\def\sbinom#1#2#3{{\textstyle\Big({\textstyle #1\atop\textstyle #2}\Big)#3^{#2}(1-#3)^{#1-#2}}}
\begin{align*}
  \sum_{\cD_n:k_i,n_i\nq}\P[\cD_n] 
  ~=~ \sum_{i_{1:n}:n_i}\P[i_{1:n}]\sum_{y_{1:n}:k_i}\P[y_{1:n}|i_{1:n}]
  ~=~ \sbinom{n}{n_i\!}{θ_i} \sbinom{n_i}{k_i}{γ_i}
\end{align*}
This is obvious or follows by explicit calculation of the sums and some algebra.
Putting everything together and finally taking the expectation over $i$ we get
\begin{align}
  \Loss_n(A|n_i,i_{n+1}=i) ~&=~ \sum_{k_i=0}^{n_i} \sbinom{n_i}{k_i}{γ_i} \ell(γ_i,k_i,n_i) \label{eq:lossni}\\
  L_i ≡ \Loss_n(A|i_{n+1}=i) ~&=~ \sum_{n_i=0}^n \sbinom{n}{n_i\!}{θ_i}~\Loss_n(A|n_i,i_{n+1}=i) \label{eq:lossi}\\
  \Loss_n(A) ~&=~ \sum_{i=1}^∞ θ_i L_i \label{eq:lossn}
\end{align}
In the deterministic case $γ_i∈\{0,1\}$ and our memorizing algorithm \eqref{eq:alg} with 0-1 loss, 
$\ell(γ_i,k_i,n_i)=[\![n_i=0]\!]$ is independent $k_i$,
so the $k_i$ sums the binomial to 1, and the $n_i$-sum collapses to $n_i=0$,
leading back to $\Loss_n(A)=\sum_{i=1}^∞θ_i(1-θ_i)^n=\EE_n=$\eqref{eq:eerrA}.

\paradot{Square loss}
For noisy labels, $γ_i∈[0;1]$,
frequency estimator $A_i=k_i/n_i$, 
square loss $\ell(γ_i,k_i,n_i)=(γ_i-k_i/n_i)^2$ for $n_i>0$ (with ``Entropy'' $γ_i(1-γ_i)$ removed),
and keeping $\ell(γ_i,k_i,0)=1$, we proceed as follows:
The $k_i$-sum in \eqref{eq:lossni} becomes the variance of $k_i/n_i$, 
hence $\Loss_n(A|n_i,i_{n+1}=i)=γ_i(1-γ_i)/n_i$.
Unfortunately plugging this into the next $n_i$-sum in \eqref{eq:lossi} leads to a hypergeometric function.
We tried various approximations, all leading essentially to the same end result.
The simplest approximation is to approximate $γ_i(1-γ_i)/n_i$ by $γ_i(1-γ_i)/(n_i+1)$, 
which is asymptotically correct and within a factor of 2 also valid for $1≤n_i<∞$, 
and avoids hypergeometric functions altogether:
\begin{align*}
  L_i ~&=~ \sum_{n_i=0}^n \sbinom{n}{n_i\!}{θ_i}~\Loss_n(A|n_i,i_{n+1}=i) \\
  ~&=~ (1-θ_i)^n ~+~ \sum_{n_i=1}^n \sbinom{n}{n_i\!}{θ_i}~\frac{γ_i(1-γ_i)}{n_i} \\
  ~&≈~ (1-θ_i)^n ~+~\sum_{n_i=1}^n \sbinom{n}{n_i\!}{θ_i}\frac{γ_i(1-γ_i)}{n_i+1} \\
  ~&=~ (1-θ_i)^n ~-~ (1-θ_i)^n γ_i(1-γ_i) ~+~ \sum_{n_i=0}^n \sbinom{n}{n_i\!}{θ_i}\frac{γ_i(1-γ_i)}{n_i+1} \\
  ~&\stackrel{(a)}=~ [1-γ_i(1-γ_i)](1-θ_i)^n ~+~ \frac{γ_i(1-γ_i)}{(n+1)θ_i} \sum_{k=1}^{n+1} \sbinom{n+1}{k}{θ_i} \\
  ~&\stackrel{(b)}=~ [1-γ_i(1-γ_i)](1-θ_i)^n ~+~\frac{γ_i(1-γ_i)}{(n+1)θ_i}[1-(1-θ_i)^{n+1}]
\end{align*}
where (a) follows from substituting $n_i=k-1$ and rearranging terms,
and (b) from adding and subtracting the missing $k=0$ contribution, and the fact that a complete binomial sums to 1.
If we assume that the noise level is the same for all features, 
i.e.\ $γ_i=γ$ or $γ_i=1-γ$, then 
\begin{align*}
  \Loss_n(A) ~&=~ \sum_{i=1}^∞ θ_i L_i 
  ~≈~ [1-γ(1-γ)]\sum_{i=1}^∞ θ_i(1-θ_i)^n ~+~ \frac{γ(1-γ)}{n+1}\sum_{i=1}^∞ [1-(1-θ_i)^{n+1}] \\
  ~&=~ [1-γ(1-γ)]\EE_n ~+~ γ(1-γ)\E[\AE_{n+1}]
\end{align*}
Again, in the deterministic case $γ∈\{0,1\}$, we get back $\Loss_n(A)=\EE_n$.
If we assume that $γ_i$ are bounded away from 0 and 1,
then still within a multiplicative constant
\begin{align*}
  \Loss_n(A) ~~\eqm~~ \EE_n ~+~ \E[\AE_n]
\end{align*}
This is quite remarkable, that the \emph{instantaneous} square loss for noisy labels 
includes a term proportional to the \emph{time-averaged} (0-1) error in the deterministic case.
As we have seen in the main paper, roughly, as long as $\EE_n$ goes to 0 slower than $1/n$, 
$\EE_n$ and $\E[\AE_n]$ have the same asymptotics,
which in turn implies that the results for deterministic classification transfer to noisy labels,
in particular $α$-Zipf-distributed data lead to $β$-power law learning curves with $β=\fr{α}{1+α}$.
While $\EE$ can decay faster than $1/n$, e.g.\ exponentially for finite models,
$\E[\AE_n]$ and hence $\Loss_n(A)$ can never decay faster than $1/n$.
As discussed in the introduction, the reason is that the accuracy to which parameters (here $γ_i$) 
can be estimated from $n_i$ i.i.d.\ data is $\eqm n_i^{-1/2}\geqm n^{-1/2}$, 
which squares to $\Loss_n(A)\geqm n^{-1}$ for (locally) quadratic loss.

\paradot{Absolute loss}
For absolute loss $\ell=|γ_i-k_i/n_i|$ there is no closed-form solution for \eqref{eq:lossni}.
For large $n_i$, the binomial is approximately Gaussian (in $k_i/n_i$) with mean $γ_i$ and variance $γ_i(1-γ_i)/n_i$,
and \eqref{eq:lossni} can be evaluated to $\sqrt{γ_i(1-γ_i)/n_i}$.
Plugging this into \eqref{eq:lossi} we can approximate the $n_i$-sum for $nθ_i\ll 1$ and for $nθ_i\gg 1$.
Plugging each into \eqref{eq:lossn}, and approximating the $i$-sum for $α$-Zipf-distributed $θ_i$
one can show that each, again, scales as $n^{-β}$ with $β=\fr{α}{1+α}$, but the latter has an additional $n^{-1/2}$ term.
For $α<1$ this is swamped by $n^{-β}$, but for $α>1$ it dominates the learning curve.
The intuition for this happening is explained in the main text.

\section{Approximating Sums by Integrals}\label{app:SumInt}

Sums $∑_{i=1}^∞ g(i)$ can be approximated by integrals $∫_1^∞ g(x)dx$.
To upper bound the approximation error classically requires computing a cumbersome integral
(Euler-Maclaurin remainder) or only works for finite sums (Trapezoid rule).
In the following we derive an upper bound on the approximation accuracy, suitable for our purpose.
First, note that for a monotone increasing function
\begin{align}\label{eq:sumintinc}
  \int_{n-1}^m g(x)dx ~≤~ \sum_{i=n}^m g(i) ~≤~ \int_n^{m+1}\!\! g(x)dx
\end{align}
with inequalities reversed for monotone decreasing functions.
Consider now any measurable function $g:[0;∞)→[0;∞)$ increasing up to $g_{max}=g(x_{max})$ and thereafter decreasing
(In our application $g(x)=f(x)e^{-nf(x)}$). Let $i_{m-1}≤x_{max}≤i_m$.
We split the integral into the increasing and decreasing part and use \eqref{eq:sumintinc} to lower-bound the error:
\begin{align*} 
  \sum_{i=1}^∞ g(i) ~&=~ \sum_{i=1}^{i_{m-1}} g(i) + \sum_{i=i_m}^∞ g(i)
  ~≥~ \int_0^{i_{m-1}}\nq g(x)dx + \int_{i_m}^∞ g(x)dx \\
  ~&=~ \int_0^∞ g(x)dx - \int_{i_m}^{i_{m-1}}\nq g(x)dx
  ~≥~ \int_0^∞ g(x)dx - g_{max}
\end{align*}
To obtain an upper bound we have to exclude $i_{m-1}$ and $i_m$ from the sums:
\begin{align*} 
  \sum_{i=1}^∞ g(i) ~&=~ \sum_{i=1}^{i_{m-2}} g(i) + \sum_{i=i_{m+1}}^∞ \!\!g(i) +~~ [g(i_{m-1}) + g(i_m)] \\
  ~&≤~ \int_1^{i_{m-1}}\nq g(x)dx + \int_{i_m}^∞ g(x)dx + ~\min\{g(i_{m-1}),g(i_m)\} + \max\{g(i_{m-1}),g(i_m)\} \\
  ~&≤~ \int_1^{i_{m-1}}\nq g(x)dx + \int_{i_m}^∞ g(x)dx + \int_{i_{m-1}}^{i_m} g(x)dx + g_{max} 
  ~=~ \int_1^∞ g(x)dx + g_{max}
\end{align*}
Together this leads to the following bound on the approximation error:
\begin{align}\label{eq:incdec}
   \bigg|∑_{i=1}^∞ g(i)-∫_a^∞ g(x)dx)\bigg| ~≤~ g_{max}
\end{align}
for any=every choice of $a∈[0;1]$. Without further assumptions on $g$, this bound is tight.
For the lower bound consider $g(x)=g_{max}$ for $i_{m-1}<x<i_m$ and $0$ otherwise.
For the upper bound consider $g(i_m)=g_{max}$ and $0$ otherwise.

\par\vspace{0pt plus \textheight}
\section{List of Notation}\label{app:Notation}
\begin{samepage}
\begin{tabbing}
  \hspace{0.13\textwidth} \= \hspace{0.73\textwidth} \= \kill
  {\bf Symbol }      \> {\bf Explanation}                                                    \\[0.5ex]
  $a/b⋅c=(a/b)⋅c$    \> ~~~~~~~~~but~~ $a/bc=a/(bc)$                                         \\[0.5ex]
  $[\![\text{Bool}]\!]$ \> 1 if Bool=True, 0 if Bool=False                                   \\[0.5ex]
  $\#\cS$            \> Number of elements in set $\cS$                                      \\[0.5ex]
  $\P,\E,\Var$       \> Probability, Expectation, Variance                                   \\[0.5ex]
  $i∈i_{1:n}$        \> is short for $i∈\{i_1,...,i_n\}$                                     \\[0.5ex]
  $~\dot=~$            \> Equal within the stated number of numerical digits                   \\[0.5ex]   
  $\eqm$             \> Equal within a multiplicative constant                               \\[0.5ex]   
  $\eqam$            \> Asymptotically or approximately proportional                         \\[0.5ex]   
  $i,j∈ℕ$            \> natural number ``feature''                                           \\[0.5ex]
  $t,n,m∈ℕ$          \> time/sample index                                                    \\[0.5ex]
  $N∈ℕ$              \> sample size                                                          \\[0.5ex]
  $θ_i$              \> probability of feature $i$                                           \\[0.5ex]
  $A$                \> tabular learning algorithm                                           \\[0.5ex]
  $h:ℕ→\cY$          \> classifier, e.g.\ binary $\cY=\{0,1\}$                               \\[0.5ex]
  $f:ℝ→ℝ$            \> Theoretical data distribution/scaling $f(i)=θ_i$                     \\[0.5ex]
  $\cD_n$            \> Data consisting of $n$ (feature,label) pairs                         \\[0.5ex]
  $\IE_n$            \> Instantaneous Error of $A$ on $i_{n+1}$ predicting $y_{n+1}$ from $\cD_n$ \\[0.5ex]
  $\EE_n$            \> Expectation of Instantaneous Error $\IE_n$ w.r.t.\ $\cD_n$           \\[0.5ex] 
  $\AE_N$            \> Time-Averaged Error from $n=1,...,N$                                 \\[0.5ex] 
  $α$                \> Exponent of Zipf distributed data frequency                          \\[0.5ex]
  $β$                \> Exponent of power law for error as a function of data size           \\[0.5ex]
  $γ$                \> Decay rate for exponential data distribution                         \\[0.5ex]
\end{tabbing}
\end{samepage}

\newpage
\section{More Figures}\label{app:Figures}

\begin{figure*}[tbh!]
\begin{center}
\includegraphics[width=0.49\textwidth]{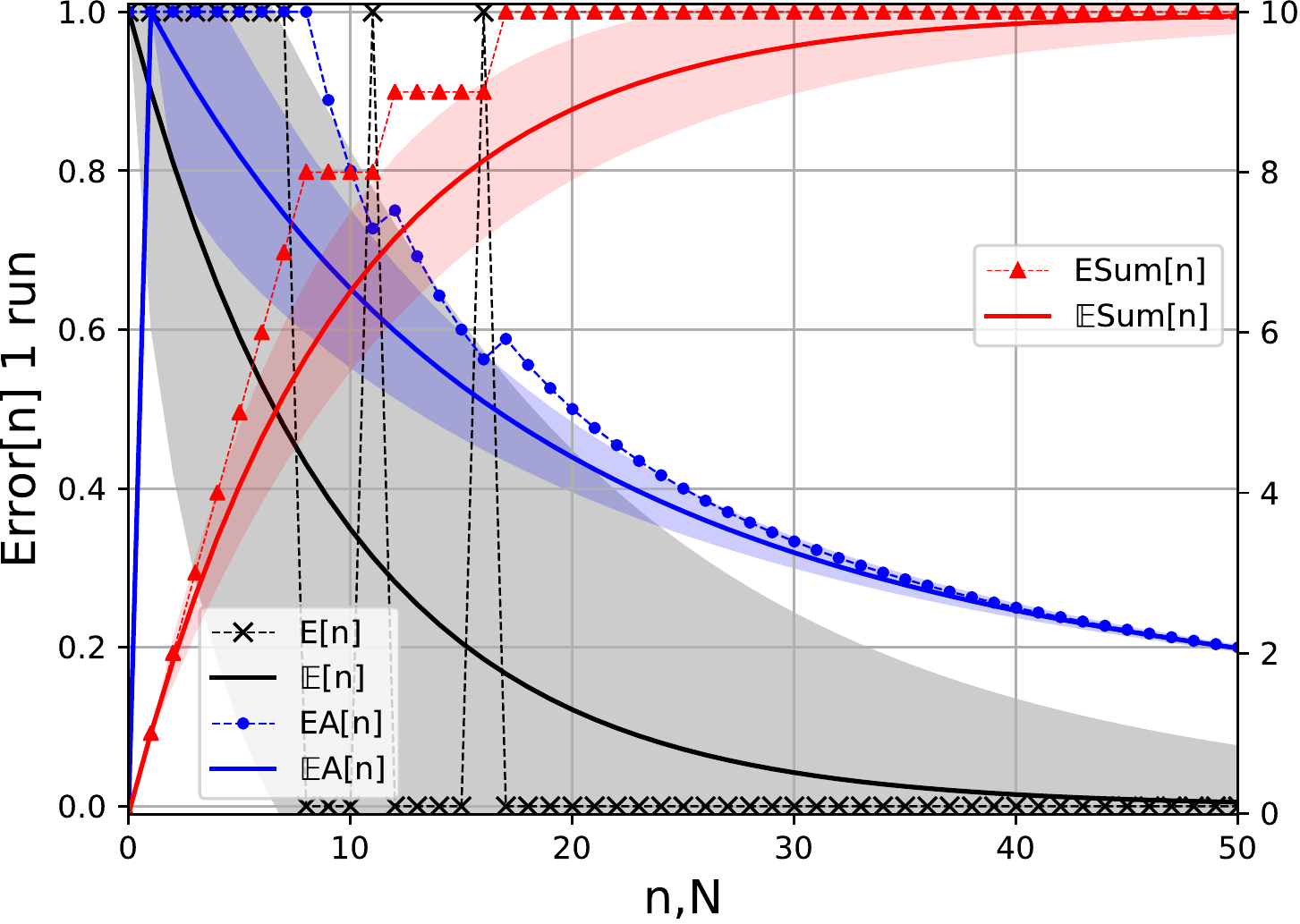}~~%
\includegraphics[width=0.49\textwidth]{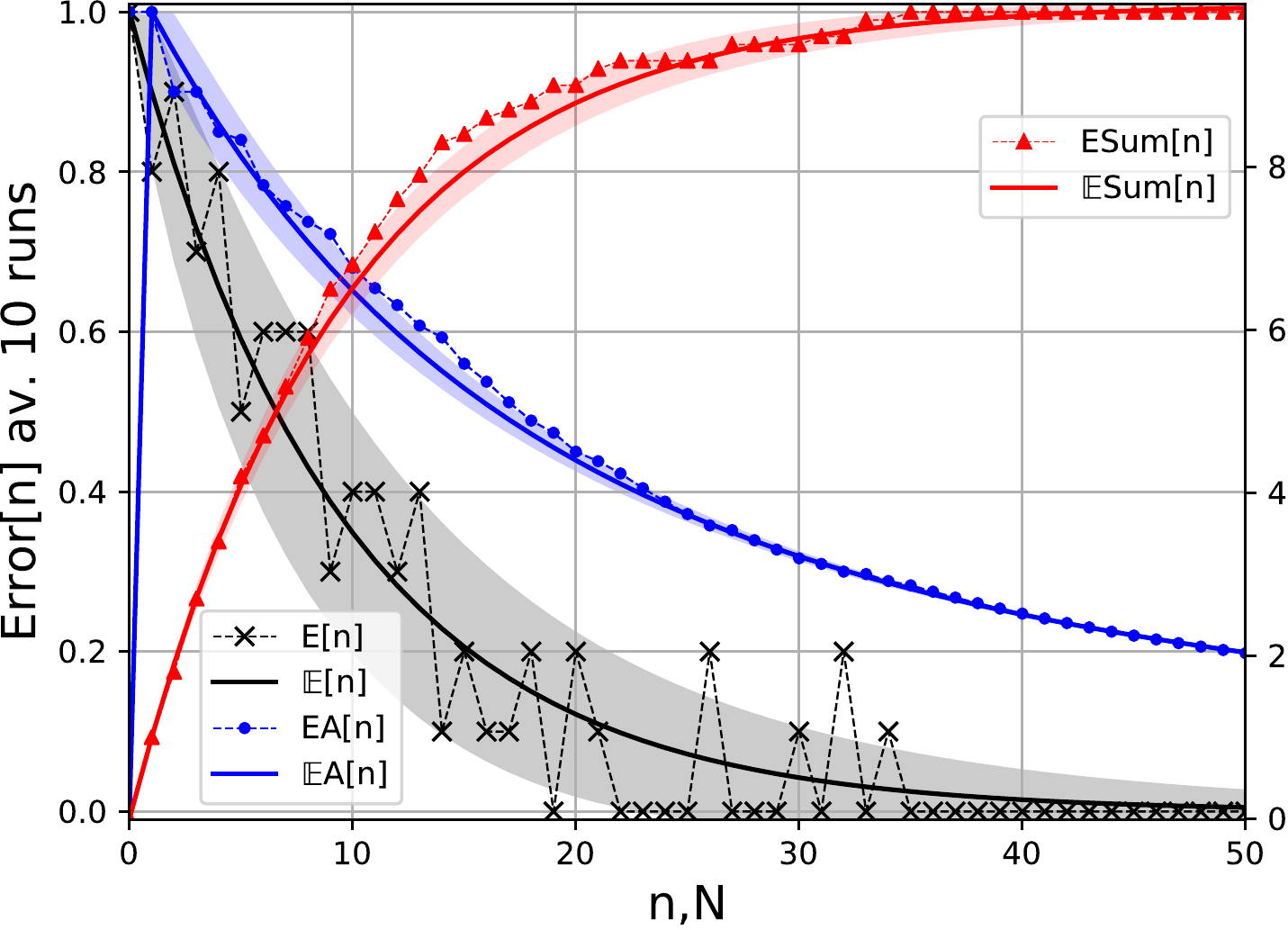} \\
\includegraphics[width=0.49\textwidth]{FigUnif100.pdf}~~%
\includegraphics[width=0.49\textwidth]{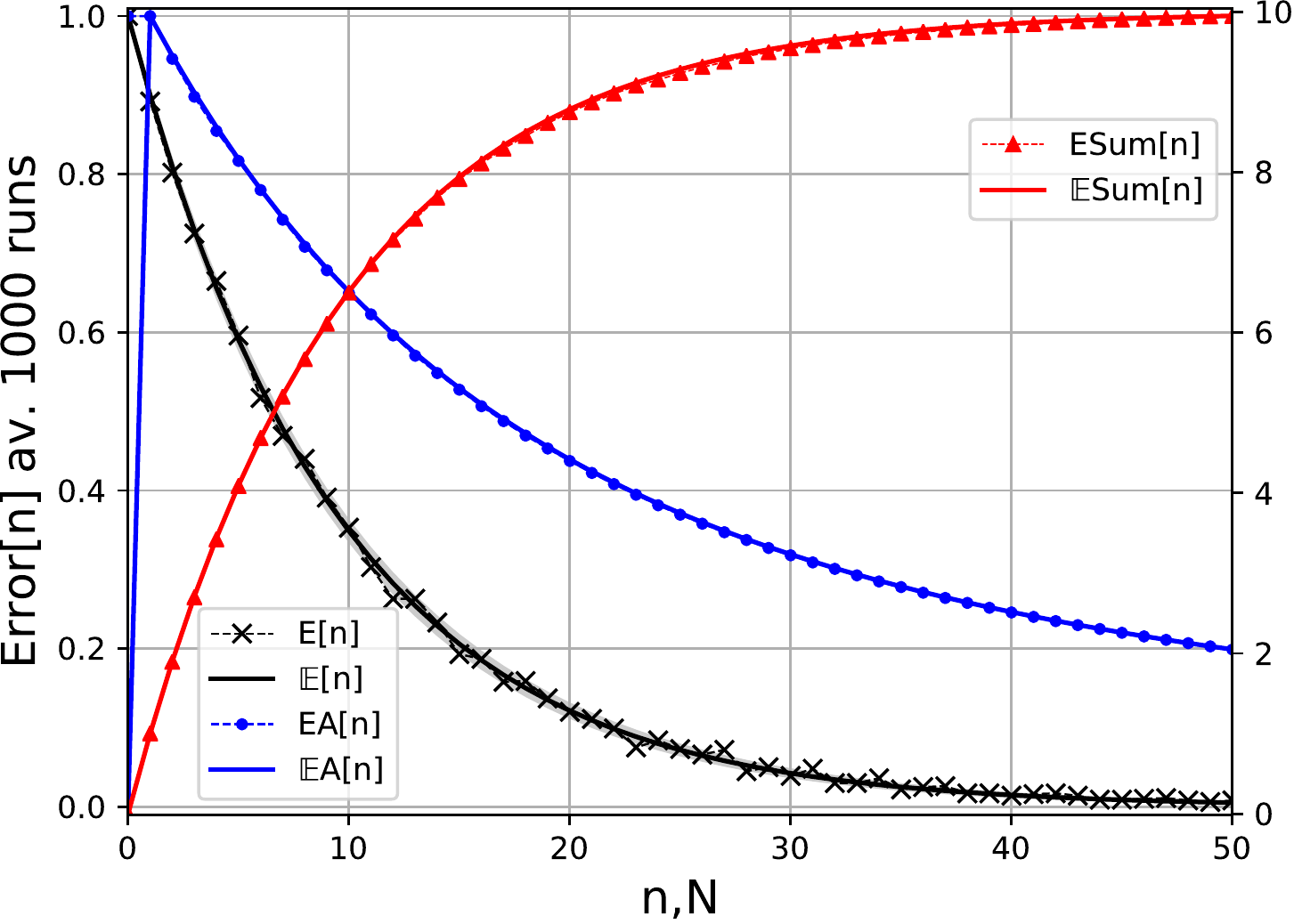}
\caption{{(\bf Learning Curves)} for uniform data distribution $\P[i_n=i]=θ_i=\fr1m$ for $i≤m=10$ averaged over $k=1,10,100,1000$ runs.
}\label{fig:unifA}\vspace*{-4ex}
\end{center}
\end{figure*}

\begin{figure*}[!tbh]
\begin{center}
\includegraphics[width=0.49\textwidth]{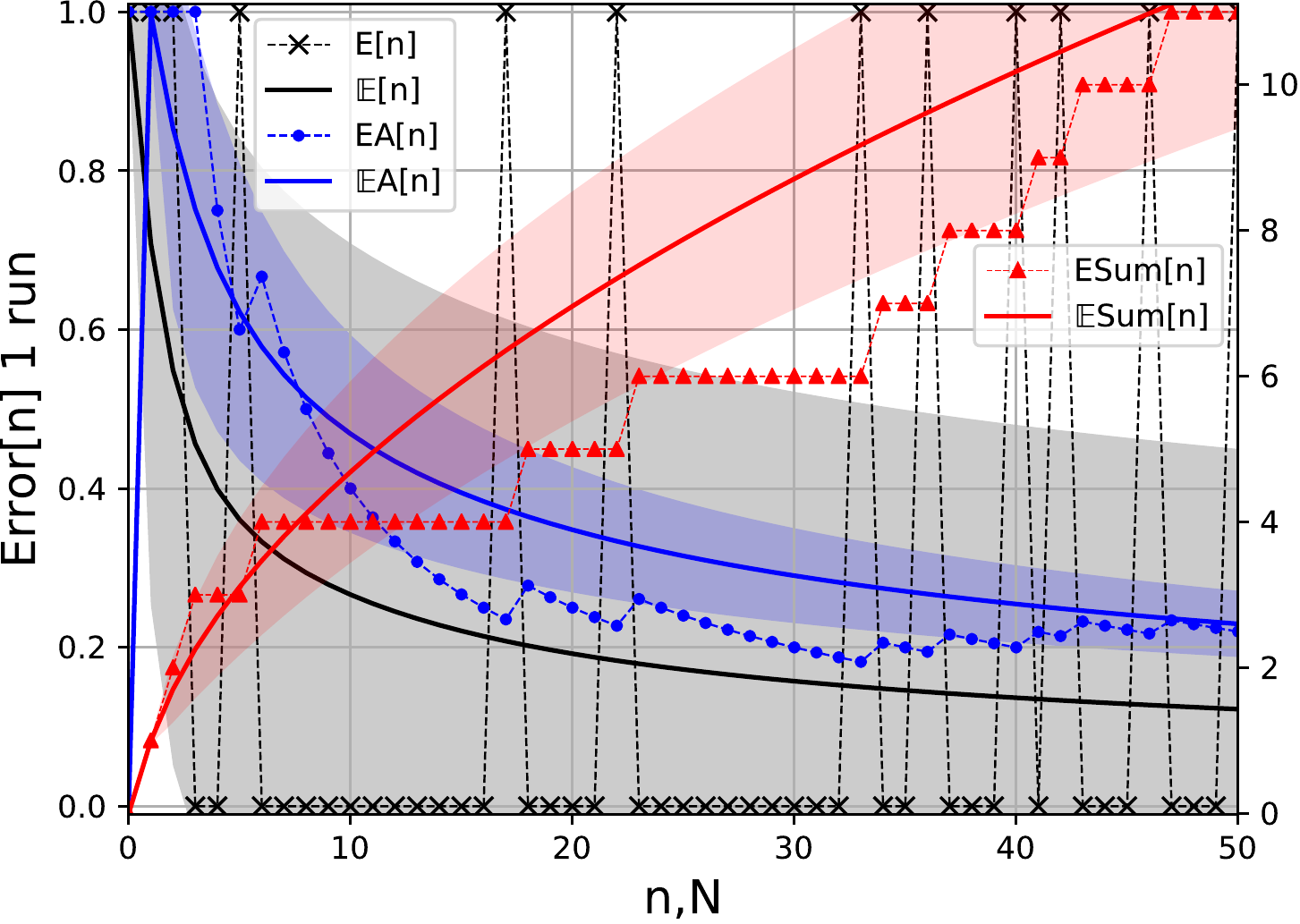}~~%
\includegraphics[width=0.49\textwidth]{FigZipf10.pdf} \\
\includegraphics[width=0.49\textwidth]{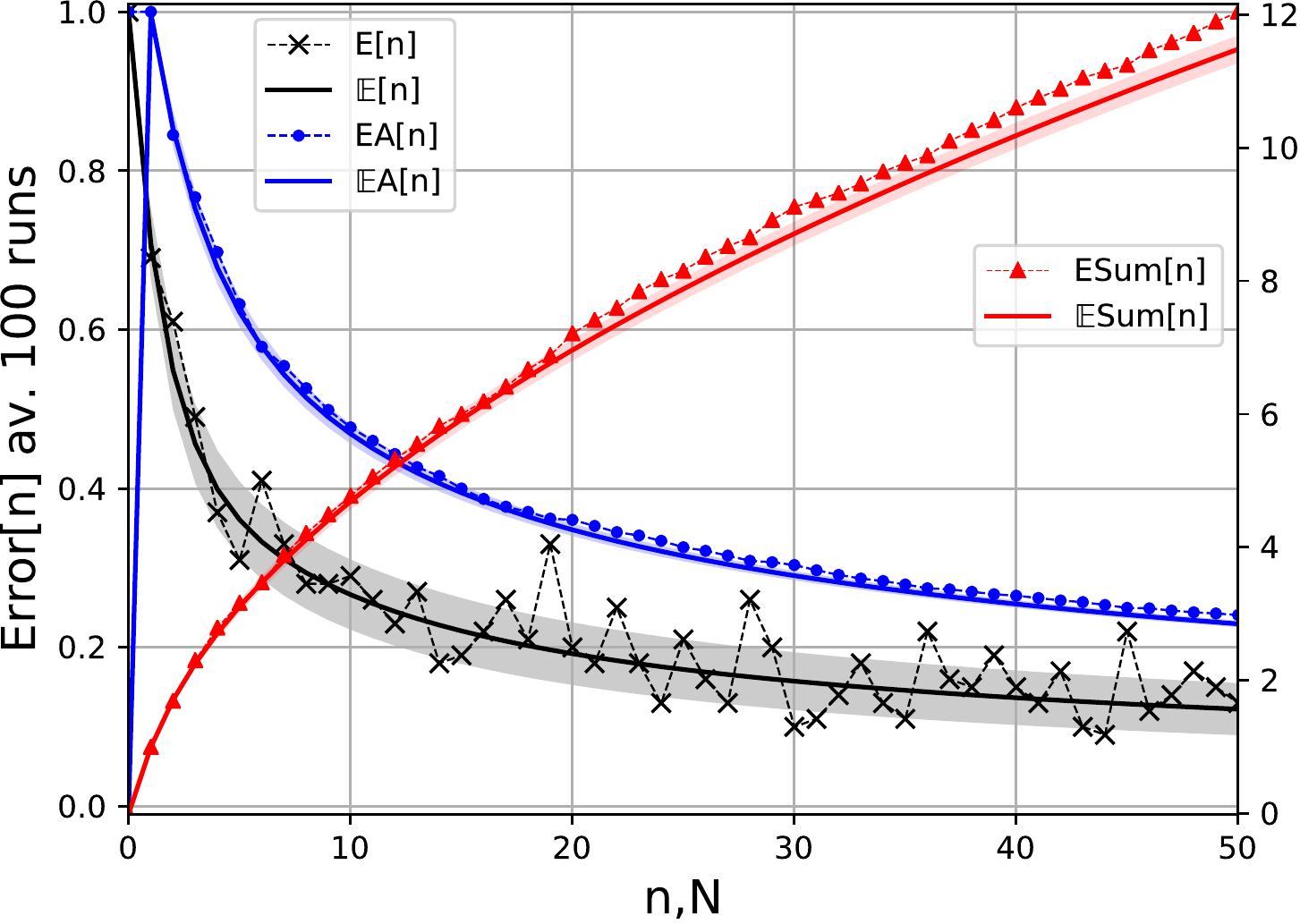}~~%
\includegraphics[width=0.49\textwidth]{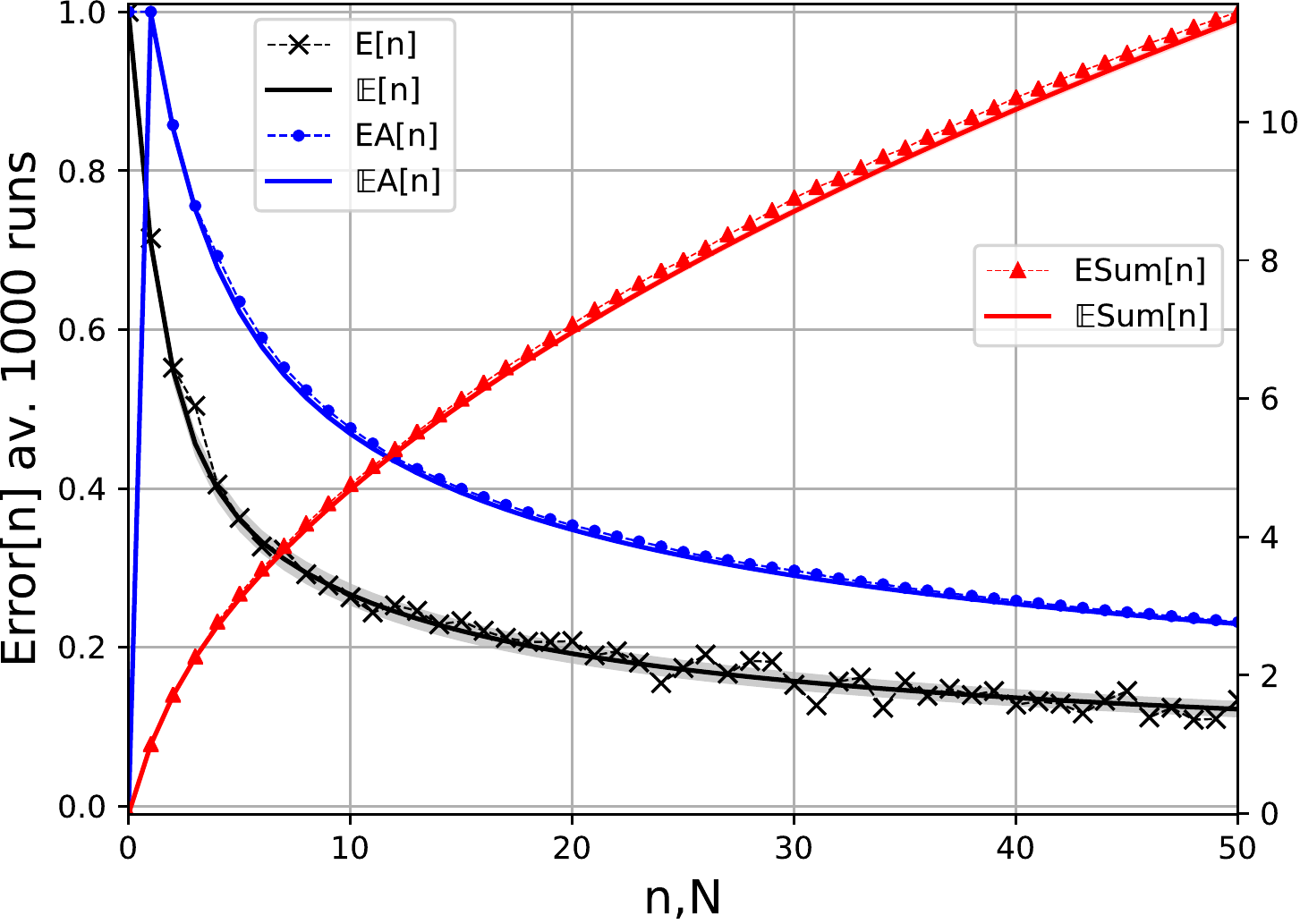}
\caption{{(\bf Learning Curves)} for Zipf-distributed data $\P[i_n=i]=θ_i\propto i^{-(α+1)}$ for $α=1$ averaged over $k=1,10,100,1000$ runs. 
}\label{fig:zipfA}\vspace*{-4ex}
\end{center}
\end{figure*}

\begin{figure*}[tbh!]
\begin{center}
\includegraphics[width=0.49\textwidth]{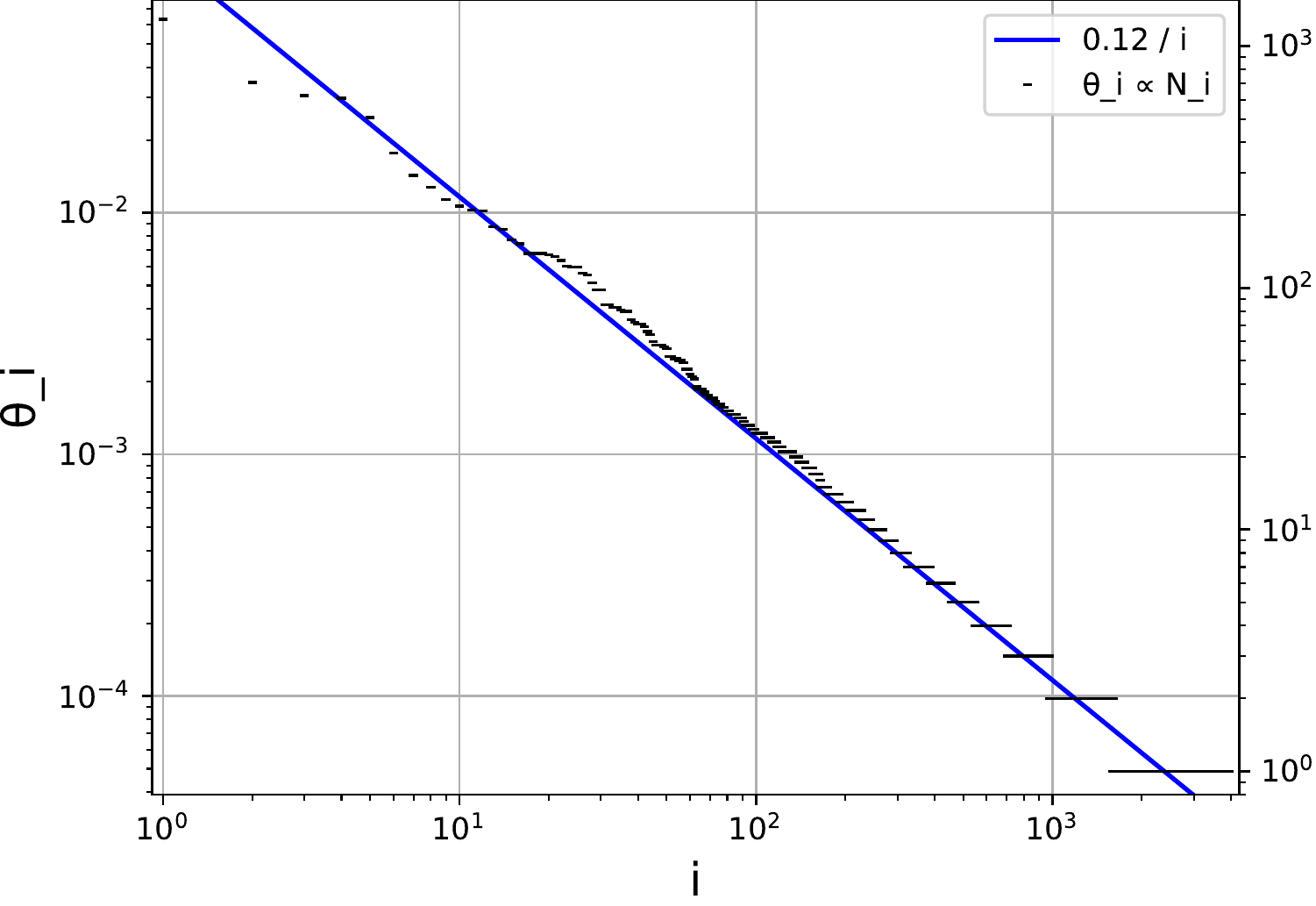}~~%
\includegraphics[width=0.49\textwidth]{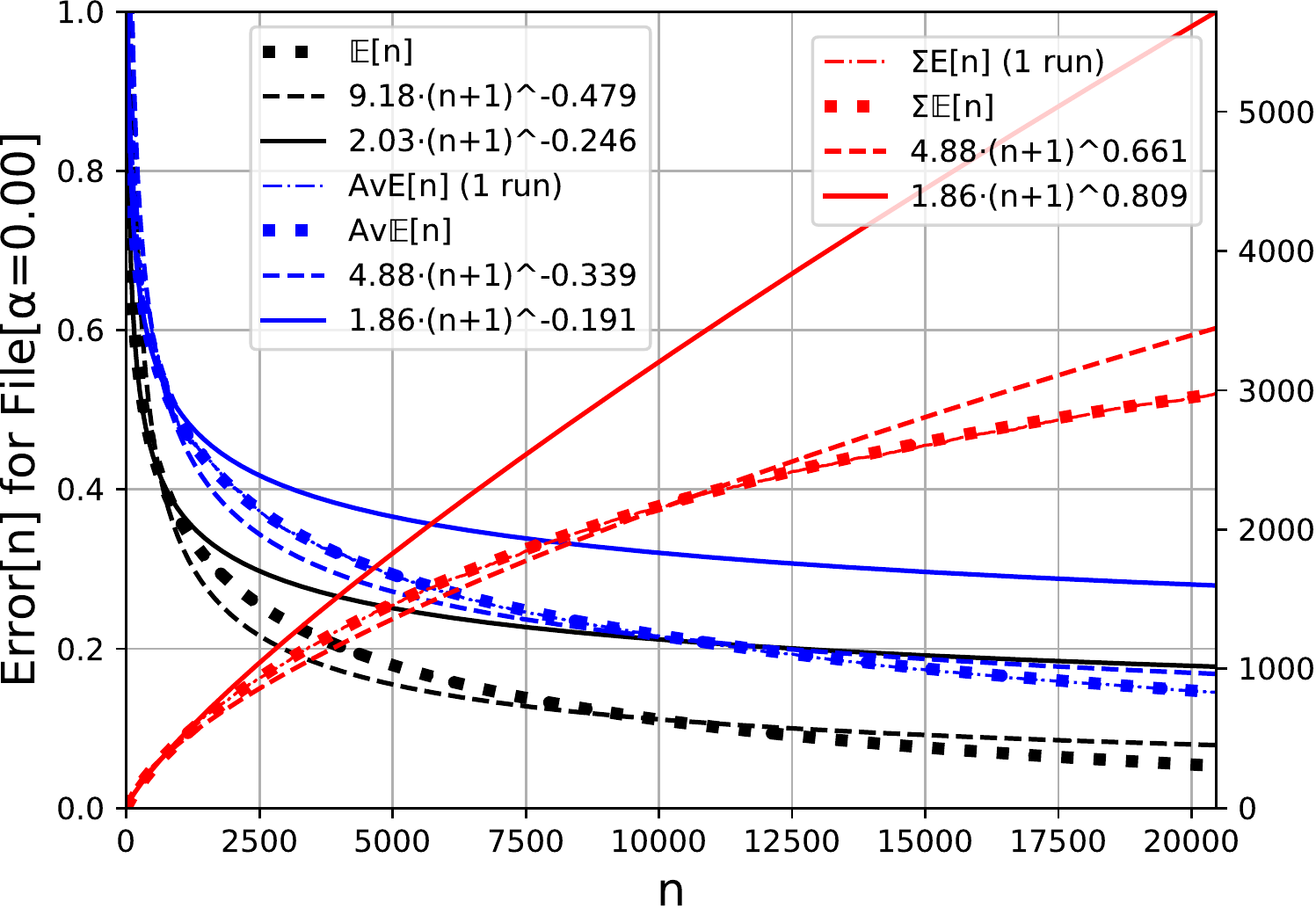}
\caption{{(\bf Word-Frequency in Text File, Learning Curve, Power Law)} \emph{(left)} 
Relative (left scale) and absolute (right scale) of frequency of words 
in the first 20469 words in file `book1' of the Calgary Corpus, 
and fitted Zipf law, which is a straight line in the Log-log plot.
\emph{(right)} Power law fit to learning curve for this data set for a word classification task. 
For large $n$, low frequency words break the Zipf law and hence the power law.
The solid line is the same as in Figure~\ref{fig:TextZipf} (right) fit to the reliable region $n≤1000$ and then extrapolated.
The dashed line is fit over the whole range of $n$.
}\label{fig:TextZipfA}\vspace*{-4ex}
\end{center}
\end{figure*}

\end{document}